\documentclass{bmvc2k}

\usepackage{amsmath,amssymb}
\usepackage{booktabs}
\usepackage{multirow}
\usepackage{graphicx}
\usepackage{xcolor}
\usepackage{array}
\usepackage{url}
\usepackage{bm}
\usepackage{xspace}
\usepackage{placeins}
\usepackage{float}
\hypersetup{hypertexnames=false}
\title{GeoT2V-Bench: Benchmarking 3D Consistency in Text-to-Video Models via 3D Reconstruction}

\addauthor{Chenrui Fan \quad Paolo Favaro}{chenrui.fan@unibe.ch, paolo.favaro@unibe.ch}{1}
\addinstitution{Computer Vision Group, Institute of Computer Science\\
University of Bern, Switzerland}

\runninghead{Fan and Favaro}{GeoT2V-Bench}

\newcommand{\bench}{\textsc{GeoT2V-Bench}\xspace}
\newcommand{\geco}{GeCo\xspace}
\newcommand{\gecoeval}{GeCo-Eval\xspace}

\begin{document}

\maketitle

\begin{abstract}
Camera-prompted text-to-video (T2V) models are increasingly used to synthesize virtual camera captures: prompts ask them to orbit objects, move through rooms, or observe static worlds from changing viewpoints. In this setting, visual plausibility is not enough; the generated frames should also provide coherent multi-view evidence for a single static 3D scene. Existing benchmarks measure visual quality, text alignment, temporal smoothness, physical plausibility, compositionality, or local geometric artifacts, but they do not directly characterize how well a generated clip supports an explicit rigid 3D reconstruction.
We introduce \bench, a reconstruction-based diagnostic benchmark for camera-prompted T2V outputs. \bench provides an answer to the question: when a generated clip depicts a static scene under camera motion, how consistent is it with an explicit rigid 3D reconstruction of that scene? Our pipeline estimates per-frame camera intrinsics and poses with VGGT-style geometry estimation~\citep{wang2025vggt}, fits DeformableGS~\citep{yang2024deformable}, derives a static MedianGS proxy by temporal-median aggregation, and renders this proxy along the estimated camera path. Rather than producing a pass/fail label or a single scalar ranking, \bench reports a continuous reconstruction profile: apparent image motion, estimated trajectory behavior, MedianGS static rendering error, static-render flow agreement, and the gap between flexible and static fits. On a fair-format four-seed evaluation with 3{,}840 completed reconstructions, covering 12 open-weight model configurations, and 80 \gecoeval static-scene prompts, the profile shows that visible motion, static rendering error, flow agreement, and flexible-vs-static behavior often disagree. In comparison with \geco~\citep{gu2025geco}, \bench captures complementary failure modes that arise when enforcing an explicit global 3D reconstruction. To support the interpretation of these reconstruction-based diagnostics, our ControlBench stress-test suite uses real static acquisitions and real-derived transformed controls to examine their behavior under frozen content, digital zoom, lateral warps, texture repainting, and foreground deformation.
\end{abstract}

\section{Introduction}
\label{sec:introduction}

Text-to-video (T2V) models are increasingly prompted with camera language:
``a slow orbit around a statue'', ``a dolly through a room'', ``a lateral pass across a street'', or ``a fly-through of a static environment.''
Such prompts ask for more than a plausible sequence of frames.
They ask the model to synthesize a camera acquisition: views of a largely static scene observed from a moving camera.
For these outputs, evaluation should ask whether the video provides coherent multiview evidence for a static 3D scene, not only whether it is sharp, realistic, or temporally smooth.

This is difficult for two reasons.
First, visually plausible videos can fail camera acquisition in different ways.
A slow-orbit prompt may produce a nearly static image, a two-dimensional zoom that imitates parallax, a motion-rich sequence with drifting texture, or a clip that can only be explained by deforming the scene over time.
These failures occur at different levels: active motion, camera-induced parallax, object permanence, global camera coherence, and static-scene explainability.
Second, reconstruction-based evaluation is itself fragile.
Camera pose estimation may fail when the video contains weak parallax, inconsistent geometry, or view-dependent repainting; even with estimated poses, static 3D reconstruction may fail because the frames do not correspond to one stable scene.
A naive reconstruction score can therefore conflate failures of the generated video with failures of the pose estimator or the reconstruction method.

\begin{figure*}[t]
\centering
\includegraphics[width=1.0\linewidth]{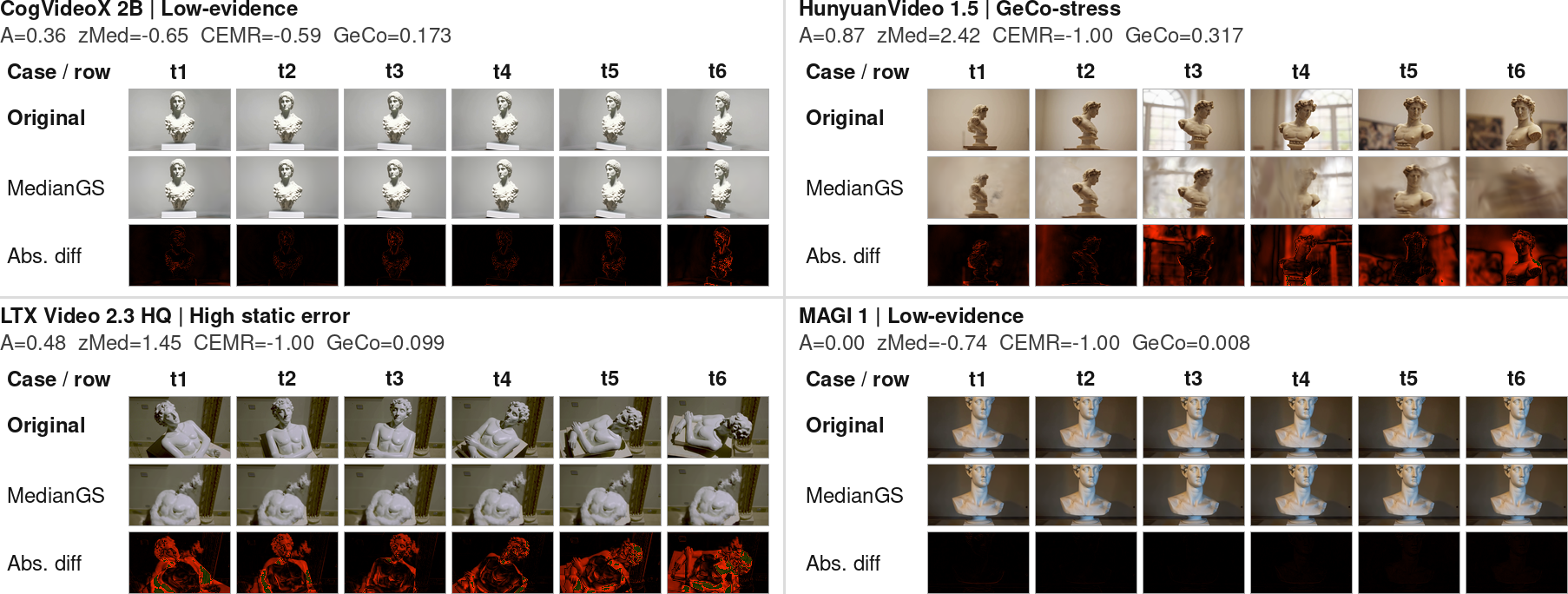}
\caption{Visual cases motivating acquisition-level evaluation.
Each case shows generated observations, MedianGS same-camera renders, and absolute differences.
The examples illustrate why static rendering error alone is insufficient: some clips are easy to render because they provide little acquisition evidence, some render individual frames while failing to explain temporal motion, and some active clips produce large static residuals or flexible-vs-static disagreement. This contrast motivates a continuous profile that separates active evidence, static reconstruction behavior, and temporal-motion explanation.}
\label{fig:teaser-acquisition-cases}
\end{figure*}

Existing evaluation methods only partially address this problem.
General video metrics emphasize perceptual quality, temporal smoothness, or distributional similarity, but do not test whether frames form a coherent camera acquisition.
Recent geometry-aware evaluation makes important progress.
In particular, \geco shows that generated videos can look photorealistic while violating local 3D geometry, and provides dense motion, structure, and inconsistency diagnostics for static-scene camera-motion prompts~\cite{gu2025geco}.
However, local geometric consistency does not establish global static-scene explainability.
A clip may appear locally consistent because it barely moves, while another may contain rich motion yet still fail to support a reliable camera trajectory or a single static 3D explanation (see Figure~\ref{fig:teaser-acquisition-cases}).
The missing test is therefore reconstructive: can the generated frames be explained as observations of one static scene along one estimated camera path?

We propose \bench, a benchmark for evaluating camera-prompted T2V outputs as synthetic static-scene acquisitions.
The benchmark is designed around the two fragile stages of reconstruction-based evaluation.
For camera estimation, it uses VGGT~\citep{wang2025vggt}, a strong feed-forward method for estimating camera intrinsics and poses from image collections.
For reconstruction, it avoids forcing a brittle static model from the outset.
Instead, it first fits an over-parameterized DeformableGS representation~\citep{yang2024deformable}, which can absorb inconsistencies as temporal variation.
We then introduce MedianGS, a static proxy obtained by temporal-median aggregation of this flexible reconstruction.
MedianGS treats the static scene as the temporally stable component of the sequence and renders it along the original VGGT-estimated cameras.

This same-camera comparison separates flexible fit from static explainability.
If the generated clip behaves like a static-scene acquisition, the MedianGS proxy should reproduce the frames under the estimated camera path.
If DeformableGS fits the video but MedianGS fails, the sequence is better explained as time-varying appearance than as a stable 3D scene.
To test whether the static proxy also explains temporal motion, we introduce CEMR, a camera-explained motion ratio that compares optical flow in the generated video with optical flow induced by the MedianGS static render.
Together, the residuals, static-render flow agreement, static-vs-flexible gap, and deformation energy identify whether failure arises from insufficient motion, unstable camera geometry, view-dependent repainting, or nonphysical scene change.

Our contributions are:
\begin{itemize}
  \item We formulate camera-prompted T2V evaluation as static-scene acquisition evaluation, shifting the focus from video plausibility to reconstructable multiview evidence.
  \item We characterize the staged difficulty of reconstruction-based evaluation: camera pose estimation and 3D reconstruction can both fail because of the inconsistencies being measured.
  \item We introduce a same-camera static-explainability test that combines VGGT-estimated camera trajectories with a MedianGS static proxy derived from a flexible DeformableGS fit.
  \item We evaluate generated videos with a continuous reconstruction profile, including MedianGS appearance, CEMR static-render flow agreement, the static-vs-flexible gap, and deformation energy, and compare these diagnostics with \geco to show where local-geometry and reconstruction-based evaluations diverge.
\end{itemize}


\section{Related Work}
\label{sec:related}

\paragraph{Video generation benchmarks.}
General video-generation benchmarks evaluate whether generated clips are visually plausible, temporally stable, and semantically aligned with prompts.
VBench and VBench++ decompose video quality into dimensions such as subject consistency, background consistency, motion smoothness, temporal flicker, aesthetic quality, and imaging quality~\cite{huang2024vbench,huang2025vbench++}.
VBench-2.0 broadens evaluation toward intrinsic faithfulness, commonsense, and physical reasoning~\cite{zheng2025vbench}.
EvalCrafter similarly aggregates visual quality, text-video alignment, motion quality, and temporal consistency~\cite{liu2024evalcrafter}, while T2V-CompBench focuses on compositional prompt satisfaction~\cite{sun2025t2v}.
These benchmarks remain important controls, but they do not directly test whether a camera-prompted clip can be used as multi-view evidence for geometry.

\paragraph{Geometry-aware generated-video metrics.}
Geometry-aware video evaluation asks a stricter question: whether generated frames obey 3D constraints.
\geco is the closest prior work.
It detects local geometric deformation by comparing observed optical flow with camera-induced rigid flow, and detects occlusion/disocclusion inconsistency through depth reprojection~\cite{gu2025geco}.
Its motion, structure, and fused scores provide dense local diagnostics for generated static-scene videos.
MEt3R and related multi-view consistency metrics evaluate whether generated views are mutually compatible, often through feature-space warping or feed-forward geometry prediction~\cite{asim2025met3r}.
WorldScore evaluates world-generation outputs with geometric consistency signals based on reprojection~\cite{duan2025worldscore}, and DynamicEval argues that dynamic-camera videos require metrics beyond ordinary temporal smoothness~\cite{babu2025dynamiceval}.
These methods are necessary for our setting, but local or pairwise consistency alone does not show that the entire clip supports a coherent full-video camera trajectory or a stable 3D representation.
\geco answers where local geometry fails; \bench asks whether the clip survives the acquisition stack.
Likewise, MEt3R-style pairwise or multi-view compatibility does not by itself establish full-video camera evidence or reconstruction utility.

\paragraph{Camera-controllable and world-generative video models.}
A parallel line of work improves camera controllability in video generation through camera trajectories, motion embeddings, 3D layout, or training-free control.
MotionCtrl, CameraCtrl, VD3D, and CamTrol are representative examples~\cite{wang2024motionctrl,he2024cameractrl,bahmani2025vd3d,hou2024training}.
These methods aim to make video generators obey camera paths.
Our goal is complementary: we audit arbitrary camera-prompted outputs, including models without explicit camera conditioning, and ask whether their generated frames behave as valid acquisition evidence.

\paragraph{Video-to-3D and Gaussian reconstruction.}
Gaussian Splatting and its deformable variants provide practical representations for reconstructing static and dynamic scenes from images or monocular video~\citep{kerbl20233d,yang2024deformable}.
Feed-forward geometry estimators such as VGGT estimate cameras, depth, and point maps from image collections, providing a practical camera-evidence backend for acquisition audits~\citep{wang2025vggt}.
Other feed-forward geometry and calibration systems, including DUSt3R, MASt3R, MapAnything, and CalibAnyView, provide natural alternatives for camera and geometry backend ablations~\citep{wang2024dust3r,leroy2024grounding,keetha2025mapanything,li2026calibanyview}.
Generative sparse-view reconstruction pipelines such as ReconX may stress generated clips differently from our discriminative geometry-and-Gaussian stack~\citep{liu2026reconx}.
Prior work has also used 3D reconstruction to inspect generated-video geometry, for example by applying 3DGS-style reconstruction to Sora videos~\citep{li2024sora}.
This is a useful precedent, but it also highlights the main difficulty: when generated videos are view-inconsistent, a direct static 3DGS reconstruction can become brittle.
In our benchmark, these methods are not introduced as new reconstruction algorithms.
They are diagnostic instruments.
DeformableGS is intentionally over-flexible: it tests whether a time-varying 4D representation can fit the generated observations.
MedianGS static reconstruction then asks how much of that fit can be converted into static-scene evidence.

\section{Generated Videos as Synthetic 3D Acquisitions}
\label{sec:acquisition}

\subsection{Problem Setup}
\label{sec:problem-setup}

We write a generated video as
\[
V=\{I_t\}_{t=1}^{T}, \qquad I_t\in\mathbb{R}^{H\times W\times 3},
\]
where $I_t$ is the RGB frame at time $t$, $T$ is the number of frames, and $H$ and $W$ are the frame height and width.
Given $V$ and a camera-motion prompt $p$, we do not assume that the clip corresponds to a real camera path.
Instead, \bench treats the clip as a candidate acquisition and asks which acquisition assumptions it satisfies.
The output is a continuous reconstruction profile rather than a scalar pass/fail label, recording apparent motion, VGGT-estimated trajectory behavior, flexible and static Gaussian rendering errors, static-render flow agreement, static-vs-flexible disagreement, and deformation diagnostics:
\[
\mathcal{P}(V,p)=
\{\bar F_{\mathrm{orig}}, \Theta_s, E_\mathrm{fit}, E_\mathrm{med}^{\mathrm{LPIPS}}, \mathrm{CEMR}, \Delta_\mathrm{stat}, E_\mathrm{deform}, s\}.
\]
Here $\mathcal{P}(V,p)$ is the reconstruction profile for video $V$ and prompt $p$.
$\bar F_{\mathrm{orig}}$ is the mean adjacent-frame optical-flow magnitude of the input video; $\Theta_s$ is the cumulative turn of the smoothed VGGT-estimated camera-center trajectory; $E_\mathrm{fit}$ is the DeformableGS flexible rendering error; $E_\mathrm{med}^{\mathrm{LPIPS}}$ is the MedianGS static rendering error measured with LPIPS; CEMR is the static-render flow-agreement score; $\Delta_\mathrm{stat}=E_\mathrm{med}^{\mathrm{LPIPS}}-E_\mathrm{fit}^{\mathrm{LPIPS}}$ is the static-vs-flexible LPIPS gap; $E_\mathrm{deform}$ is a deformation-energy diagnostic; and $s$ stores completion flags and qualitative diagnostic notes.
\geco is used later as an external local-geometry comparison.

\subsection{Acquisition Assumptions}
\label{sec:assumptions}

We define five assumptions for camera-prompted static-scene videos.

\paragraph{Video-evidence assumption.}
The clip should contain active observations: nontrivial viewpoint or scene evidence, sufficient texture/detail, and stable temporal behavior.
A near-static clip may be easy to reconstruct but does not satisfy a camera-motion prompt.

\paragraph{Local-geometry assumption.}
Co-visible surfaces should move consistently with camera-induced parallax, and occlusion/disocclusion events should preserve object permanence.
We later compare against \geco as an external diagnostic for this local-geometry layer.

\paragraph{Camera assumption.}
The frames should induce a non-degenerate VGGT-estimated camera trajectory.
A generated video may contain large apparent motion while failing to induce a coherent camera path.

\paragraph{Static-reconstruction assumption.}
For nominally static prompts, the observations should be reproducible by a static MedianGS proxy rendered along the VGGT-estimated intrinsics and poses.
If only a deformable representation can fit the frames, the clip may rely on view-dependent repainting or nonphysical drift.

\paragraph{Deformation-discipline assumption.}
When a prompt describes a nominally static scene, learned deformation should not become the primary explanation for the observations.
High deformation energy is not automatically a failure, but high deformation together with a large static reconstruction gap is an outlier-level warning that the clip may require nonphysical time-varying geometry.

\subsection{Why Reconstruction Residual Is Not Video Quality}
\label{sec:residual-not-quality}

A central design choice is that reconstruction residual is interpreted as acquisition stress, not visual quality.
A low residual can indicate static-scene evidence, but it can also indicate low motion or low scene complexity.
A high residual can indicate geometric inconsistency, but it can also arise from high motion, repeated texture, occlusion, specular appearance, or limitations of the reconstruction backend.
Therefore, \bench reports MedianGS rendering error together with apparent flow, VGGT-estimated trajectory diagnostics, CEMR, the static-vs-flexible gap, and deformation diagnostics; \geco is reported separately as a local-geometry comparison.
The correct unit of interpretation is the acquisition profile, not a scalar rank.

\section{GeoT2V-Bench}
\label{sec:benchmark}

\subsection{Prompt and Model Substrate}
\label{sec:substrate}

We use the 80 controlled static-scene prompts from \gecoeval as the acquisition substrate~\citep{gu2025geco}.
The prompts cover object-centric, indoor, outdoor, and appearance-complex cases, and are designed to stress camera motion, occlusion, repeated structure, texture, reflection, and object permanence.
We use this prompt suite because it is already a controlled static-scene camera-motion stress substrate, allowing us to isolate acquisition-level questions without introducing a new prompt distribution.
The main reported audit uses fair-format reconstructions for 3840 generated clips: 12 open-weight configurations, 80 prompts, and four generation seeds $\{0,1,2,3\}$.
This reported instance focuses on open-weight configurations with locally available videos.
The panel covers CogVideoX~\citep{yang2025cogvideox}, HunyuanVideo~\citep{kong2024hunyuanvideo,wu2025hunyuanvideo}, LTX-Video/LTX-2~\citep{hacohen2024ltx,hacohen2026ltx}, MAGI-1~\citep{teng2025magi}, Open-Sora~\citep{zheng2025open}, SkyReels~\citep{chen2025skyreels}, and Wan~\citep{wan2025wan} variants.
The prompt categories are balanced across object-centric, indoor, outdoor, and appearance-stress cases (20 each), with 40 slow and 40 fast camera-motion prompts.

\subsection{Pipeline Overview}
\label{sec:pipeline}

A camera-motion prompt and generated frames are treated as a candidate acquisition.
The audit first records whether the video contains visible, coherent acquisition evidence, then measures how well the same frames can be reproduced by a static Gaussian proxy under the estimated camera intrinsics and poses.

\begin{figure*}[t]
\centering
\includegraphics[width=\linewidth]{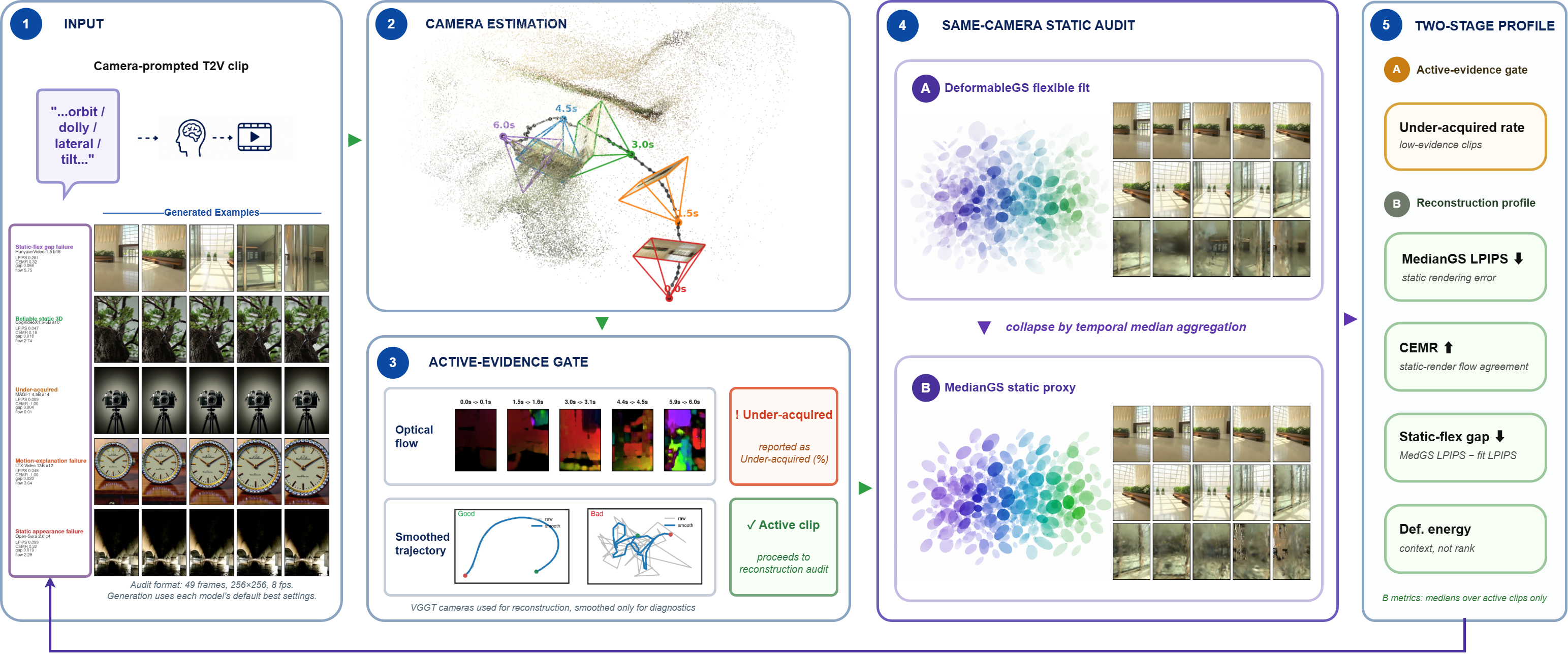}
\caption{Reconstruction-profile pipeline.
The five generated clips in the input panel are carried through the same audit and receive the example-level outcomes shown beside them, spanning under-acquisition, reliable static 3D, static-appearance failure, motion-explanation failure, and static-flex gap failure.
\bench first records active acquisition evidence using apparent flow and smoothed VGGT trajectory-coherence descriptors.
DeformableGS and MedianGS then use the original VGGT-estimated camera intrinsics and poses.
\bench derives a MedianGS static proxy from the DeformableGS fit and reports static appearance error, static-render flow agreement, and the static-vs-flexible gap as continuous diagnostics.
Deformation energy, backend checks, and the \geco comparison provide additional context in the experiments.}
\label{fig:pipeline}
\end{figure*}

\subsection{\texorpdfstring{Camera Estimation and Active Acquisition Evidence}{Camera Estimation and Active Acquisition Evidence}}
\label{sec:video-evidence}
\label{sec:camera-evidence}

Before interpreting any reconstruction residual, \bench first asks whether the video contains an acquisition attempt.
We estimate a camera trajectory $\{C_t\}_{t=1}^{T}$ from the generated frames using VGGT-style geometry estimation~\citep{wang2025vggt}, with $C_t=(K_t,\mathbf{T}_t)$.
Here ``cameras'' means the per-frame intrinsics $K_t$ and camera poses $\mathbf{T}_t$ estimated by VGGT.
All reconstructions use these original VGGT-estimated intrinsics and poses; the smoothing described below is only a descriptor for deciding whether a plausible low-frequency camera path exists.
Apparent image motion is measured by the mean adjacent-frame optical-flow magnitude $\bar F_{\mathrm{orig}}$ at the fair-format resolution.
Camera coherence is measured from the VGGT-estimated camera centers after descriptor-only centering, robust normalization, and local Savitzky--Golay smoothing~\citep{savitzky1964smoothing}.
From the smoothed diagnostic trajectory we compute a 10-point arc-length-resampled cumulative turn $\Theta_s$.
We also report smoothed path length, raw-to-smoothed jitter $\rho_j$, and per-frame FOV variation as camera-instability context, because VGGT center jitter can be coupled with focal-length changes while the original VGGT-estimated intrinsics and poses still render coherently.
	For visualization, the 3D camera centers are projected onto the first two PCA directions of the smoothed trajectory; the plot axes are relative trajectory coordinates rather than image pixels or metric world units.
	These quantities summarize active-acquisition evidence before reconstruction quality is interpreted.
	$\bar F_{\mathrm{orig}}$ says how much apparent frame-to-frame motion exists, while $\Theta_s$ asks whether the low-frequency camera path repeatedly changes direction.
	For the short fair-format clips used here, a smoothed cumulative turn much larger than one full revolution indicates an irregular trajectory rather than a plausible single camera sweep.
	Because VGGT scale is arbitrary, the trajectory descriptors are calibrated within the frozen fair-format pipeline instead of physical motion units.
	The normalized and smoothed trajectory is used for descriptors and visualization; DeformableGS and MedianGS still use the original VGGT-estimated intrinsics and poses.
	Supplementary Table~A5 lists the active-evidence thresholds, boolean rule, and trajectory-descriptor settings.
	Supplementary Figure~A1 visualizes a real acquisition reference with its frozen counterpart, plus active/inactive generated examples from orbit, dolly, lateral, roll, and tilt prompt families, together with per-frame FOV and seven sampled video frames.
	Supplementary Table~A8 adds a minimal deterministic prompt-family trajectory check for orbit, dolly, lateral, roll, and tilt prompts.
	The active-evidence gate is not a camera-command compliance score.
	It only separates clips with insufficient acquisition evidence from clips for which same-camera reconstruction diagnostics are meaningful; prompt-family trajectory compliance is reported separately as a lightweight diagnostic in Supplementary Table~A8.

\subsection{\texorpdfstring{Same-Camera Reconstruction Audit}{Same-Camera Reconstruction Audit}}
\label{sec:reconstruction-audit}

\paragraph{DeformableGS flexible fit.}
\label{sec:deformable-fitting}

Given the original VGGT-estimated intrinsics and poses, we first optimize DeformableGS~\citep{yang2024deformable} as an intentionally flexible diagnostic fit to the generated clip.
Supplementary Table~A6 lists the fixed implementation settings used by this reconstruction and metric stack.
Let $\mathcal{G}$ be the canonical Gaussian set.
For Gaussian $j$ with canonical position $\boldsymbol{\mu}_j$, rotation $\mathbf{q}_j$, and scale $\mathbf{s}_j$, and for normalized time $\tau_t$, we write the deformation field's time-varying offsets as
\[
D_\theta(\boldsymbol{\mu}_j,\tau_t)
=
(\Delta \boldsymbol{\mu}_{jt},\Delta \mathbf{q}_{jt},\Delta \mathbf{s}_{jt}).
\]
Here $D_\theta$ is the learned deformation field with parameters $\theta$, and $\Delta \boldsymbol{\mu}_{jt}$, $\Delta \mathbf{q}_{jt}$, and $\Delta \mathbf{s}_{jt}$ are the position, rotation, and scale offsets for Gaussian $j$ at frame $t$.
The flexible same-camera render is
\[
\tilde I_t^{\mathrm{def}}=\mathcal{R}(\mathcal{G},D_\theta,C_t,\tau_t),
\]
Here $\tilde I_t^{\mathrm{def}}$ is the DeformableGS render, $\mathcal{R}$ is the Gaussian renderer, and $C_t=(K_t,\mathbf{T}_t)$ denotes the VGGT-estimated camera intrinsics $K_t$ and pose $\mathbf{T}_t$ for frame $t$.
The backend optimizes the observed-frame objective
\[
\min_{\mathcal{G},\theta}\sum_{t=1}^{T}
\ell\!\left(\mathcal{R}(\mathcal{G},D_\theta,C_t,\tau_t), I_t\right).
\]
Here $I_t$ is the generated RGB frame, $T$ is the number of frames, $\ell$ is the rendering loss used by the DeformableGS optimizer, and the minimization is over the Gaussian set $\mathcal{G}$ and deformation parameters $\theta$.
We use $E$ for render-to-video error: it measures how different a rendered frame sequence is from the generated video under a specified distance $d$.
The DeformableGS flexible rendering error
\[
E_{\mathrm{fit}}=\frac{1}{|\mathcal{S}|}\sum_{t\in\mathcal{S}}
d(I_t,\tilde I_t^{\mathrm{def}})
\]
measures flexible fit under the VGGT-estimated camera path.
Here $\mathcal{S}$ is the set of evaluation frames and $d$ denotes the distance used for a specific error, such as LPIPS.
The following MedianGS step tests how much of this flexible fit can be expressed as a static scene, since a deformable representation can absorb view-dependent repainting or nonphysical drift.

\paragraph{MedianGS static proxy.}
\label{sec:mediangs}

We then derive a MedianGS static proxy $\mathcal{G}_{\mathrm{med}}$ from the time-varying DeformableGS fit by temporal-median aggregation.
For each Gaussian $j$, we aggregate the time-varying position, rotation, and scale offsets as
\[
\widetilde{\Delta\boldsymbol{\mu}}_j=\mathrm{median}_{t}\Delta\boldsymbol{\mu}_{jt},\quad
\widetilde{\Delta\mathbf{q}}_j=\mathrm{median}_{t}\Delta\mathbf{q}_{jt},\quad
\widetilde{\Delta\mathbf{s}}_j=\mathrm{median}_{t}\Delta\mathbf{s}_{jt}.
\]
Here $\mathrm{median}_{t}$ aggregates over frames, and the tilded quantities define $\mathcal{G}_{\mathrm{med}}$ by applying median offsets to the canonical Gaussians.
For rotation parameters, we follow the additive quaternion-parameter convention used by DeformableGS: the learned 4D rotation offsets are aggregated by a component-wise temporal median, added to the normalized base Gaussian rotation parameter, and renormalized before rendering the MedianGS proxy.
The static same-camera render is
\[
\hat I_t^{\mathrm{med}}=\mathcal{R}(\mathcal{G}_{\mathrm{med}},C_t).
\]
Here $\hat I_t^{\mathrm{med}}$ is the MedianGS static render at frame $t$, using the same original $C_t$ as the flexible fit.
The MedianGS static rendering error
\[
E_{\mathrm{med}}=\frac{1}{|\mathcal{S}|}\sum_{t\in\mathcal{S}}
d(I_t,\hat I_t^{\mathrm{med}})
\]
uses the same $\mathcal{S}$ and $d$ definitions as $E_{\mathrm{fit}}$ and tests whether the generated observations survive as a single static scene under the VGGT-estimated camera path.
MedianGS is not intended as the only possible static reconstruction backend.
It is a pragmatic same-camera static proxy derived from the same flexible fit, which makes the static-vs-flexible gap well defined under the same cameras, initialization history, and observed frames.
Direct static 3DGS, mean-collapse, and alternative geometry backends are therefore treated as backend sanity checks rather than separate benchmark definitions.

\subsection{\texorpdfstring{Static Explainability Metrics}{Static Explainability Metrics}}
\label{sec:gap}

\paragraph{MedianGS LPIPS.}

The first static-explainability metric is raw MedianGS LPIPS, $E_\mathrm{med}^{\mathrm{LPIPS}}$, computed between each generated frame and the corresponding MedianGS same-camera render~\citep{zhang2018unreasonable}.
Here $E_\mathrm{med}^{\mathrm{LPIPS}}$ is $E_{\mathrm{med}}$ with $d$ set to LPIPS.
This asks whether the static proxy can reproduce the observed frames.

\paragraph{Static-render flow agreement / CEMR.}

The second metric asks whether the MedianGS static render reproduces the generated clip's temporal image motion.
We call this static-render flow agreement CEMR, for camera-explained motion ratio: it measures how much of the generated clip's adjacent-frame optical flow is reproduced by rendering the static MedianGS proxy along the same VGGT-estimated intrinsics and poses.
For sampled adjacent-frame pair $i$, we compute
\[
\mathrm{CEMR}_i
= 1-\frac{\lVert F^\mathrm{orig}_i-F^\mathrm{med}_i\rVert}
{\lVert F^\mathrm{orig}_i\rVert+\epsilon}.
\]
Here $i$ indexes a sampled adjacent-frame pair, $F^\mathrm{orig}_i$ is the optical-flow field in the generated clip, $F^\mathrm{med}_i$ is the optical-flow field in the MedianGS same-camera render sequence, $\lVert\cdot\rVert$ denotes mean flow magnitude over pixels, and $\epsilon=10^{-6}$ prevents division by zero.
In the numerator, the norm is applied to the residual flow vector field before spatial averaging; it is an endpoint-style residual magnitude rather than a difference of scalar flow magnitudes.
Values near one mean the static render produces nearly the same optical flow as the generated clip; values near or below zero mean the generated temporal change is not explained by the same-camera static render.
We compute the primary CEMR with RAFT-Large optical flow~\citep{teed2020raft} on three uniformly sampled adjacent frame pairs at the fair-format $256\times256$ resolution--for 49-frame clips, pairs $(0,1)$, $(23,24)$, and $(47,48)$ in zero-based indexing--clip the averaged value to $[-1,1]$, and treat higher values as better.
Because CEMR normalizes by the generated-video flow magnitude, we interpret it jointly with the apparent-flow descriptor.
In very low-flow clips, negative or saturated CEMR is a low-evidence / motion-explanation warning.

\paragraph{Static-vs-flexible LPIPS gap.}

The third metric compares flexible fit and static-reconstruction explainability and is defined as
\[
\Delta_{\mathrm{stat}}=
E_{\mathrm{med}}^{\mathrm{LPIPS}}-
E_{\mathrm{fit}}^{\mathrm{LPIPS}} .
\]
Here $E_{\mathrm{fit}}^{\mathrm{LPIPS}}$ is the DeformableGS flexible-fit residual computed with LPIPS, and $\Delta_{\mathrm{stat}}$ measures how much worse the static MedianGS proxy is than the flexible fit.
A high gap indicates that the flexible DeformableGS fit explains the video substantially better than the static MedianGS proxy.

\paragraph{Deformation energy context.}

We also report deformation energy as context for the static-vs-flexible gap, using
\[
E_\mathrm{deform}=
\frac{1}{|\mathcal{T}||\mathcal{J}|}
\sum_{\tau\in\mathcal{T}}\sum_{j\in\mathcal{J}}
\lVert \Delta\boldsymbol{\mu}_{j}(\tau)\rVert_2 .
\]
Here $\mathcal{T}$ is the sampled set of normalized time coordinates, $\mathcal{J}$ is the evaluated Gaussian set, $\Delta\boldsymbol{\mu}_{j}(\tau)$ is the position-offset component predicted by $D_\theta$ for Gaussian $j$ at time $\tau$, and $\lVert\cdot\rVert_2$ is the Euclidean norm.
Raw deformation energy is affected by VGGT scale, scene extent, visibility, opacity, and floating Gaussians.
High static-vs-flexible gap together with high deformation energy suggests candidate nonphysical deformation, while many hard cases raise both DeformableGS and MedianGS residuals.

\subsection{\texorpdfstring{Continuous Reconstruction Profile}{Continuous Reconstruction Profile}}
\label{sec:continuous-reconstruction-profile}

The benchmark output is a continuous reconstruction profile.
For each clip we report the active-acquisition descriptors from Section~\ref{sec:video-evidence}, the static rendering error $E_\mathrm{med}^{\mathrm{LPIPS}}$, the static-render flow agreement CEMR, and the static-vs-flexible LPIPS gap $\Delta_{\mathrm{stat}}$.
At model level, we first report how many of the 320 clips fail the active-evidence check, then summarize MedianGS LPIPS, CEMR, static-vs-flexible gap, and deformation energy over the active clips only; Supplementary Table~A3 retains the unfiltered full-set medians.
Low MedianGS LPIPS indicates that the static proxy reproduces frames well, high CEMR indicates that the rendered static proxy reproduces temporal image motion, and a low static-vs-flexible gap indicates that the flexible DeformableGS fit is not substantially easier than the static proxy.
Deformation energy contextualizes the static-vs-flexible gap, and \geco is reserved for the experiment section as an external local-geometry comparison.
This presentation keeps the method tied to what is actually measured: reconstruction quality and reconstruction disagreement under the same estimated camera path.

\subsection{\texorpdfstring{Backend and Control Diagnostics}{Backend and Control Diagnostics}}
\label{sec:diagnostic-layers}

\paragraph{Backend sanity.}

Supplementary Table~A9 reports a 192-clip MASt3R feed-forward geometry cross-check; Supplementary Table~A10 and Supplementary Figure~A2 report a 100-clip COLMAP/BA sanity check; Supplementary Table~A11 and Supplementary Figure~A3 compare MedianGS with mean-collapse; Supplementary Figure~A4 shows that image-space static proxies lack the same-camera Gaussian structure of MedianGS; Supplementary Figure~A5 gives a direct vanilla static-3DGS baseline on Prompt A1 alongside DeformableGS and MedianGS; and Supplementary Table~A13 plus Supplementary Figure~A6 report the Farneb{\"a}ck and GMFlow CEMR sensitivity checks~\citep{farneback2003two,xu2022gmflow}.
Together, these experiments test whether the reconstruction profile is stable under alternative geometry, static proxy, and optical-flow choices.

\paragraph{ControlBench role.}

ControlBench is a stress-test suite for metric behavior.
It contains public real static acquisitions and real-derived transformed controls, including frozen videos, digital zoom, lateral warp, foreground warp, and texture repaint variants.
We use it to inspect whether the reconstruction diagnostics respond in plausible ways: frozen controls should have near-zero apparent motion, 2D zoom and lateral warp should stress motion and reconstruction agreement, texture repaint should raise appearance-related errors, and foreground warp should expose deformation-like behavior.
Supplementary Table~A4 reports the resulting metric profile, and Supplementary Figure~A1 visualizes how VGGT camera and FOV diagnostics behave on real, frozen, and generated examples.

\begin{table}[t]
\centering
\scriptsize
\resizebox{\linewidth}{!}{%
\begin{tabular}{lrrrrrr}
\toprule
Model & Under-acq. (\%) $\downarrow$ & Active $n$ & MedGS LPIPS $\downarrow$ & CEMR $\uparrow$ & SF gap $\downarrow$ & Def. energy \\
\midrule
CogVideoX-2B & 14.4 & 274 & 0.094 & 0.15 & 0.040 & 0.103 \\
CogVideoX-5B & 20.6 & 254 & 0.102 & 0.19 & 0.044 & 0.096 \\
CogVideoX1.5-5B & 14.7 & 273 & 0.085 & -0.24 & 0.040 & 0.093 \\
HunyuanVideo & 38.8 & 196 & 0.071 & -0.47 & 0.030 & 0.106 \\
HunyuanVideo-1.5 & 11.6 & 283 & 0.327 & 0.08 & 0.056 & 0.194 \\
LTX-2.3 dev/HQ & 24.4 & 242 & 0.181 & 0.21 & 0.064 & 0.139 \\
LTX-2.3 dev/default & 29.4 & 226 & 0.198 & 0.09 & 0.049 & 0.139 \\
LTX-Video 13B & 21.2 & 252 & 0.075 & -0.13 & 0.034 & 0.084 \\
MAGI-1 4.5B & 45.0 & 176 & 0.054 & -1.00 & 0.033 & 0.056 \\
Open-Sora 2.0 & 28.1 & 230 & 0.115 & -0.04 & 0.043 & 0.087 \\
SkyReels-V2 & 16.2 & 268 & 0.065 & -0.14 & 0.029 & 0.090 \\
WAN 2.2 & 31.6 & 219 & 0.138 & 0.21 & 0.044 & 0.115 \\
\bottomrule
\end{tabular}}
\caption{Model-level two-stage acquisition and reconstruction profile on the fair-format four-seed audit. Each model contains 320 generated clips from 80 prompts and four seeds. Under-acq. (\%) reports the fraction of clips that fail the active-evidence gate before reconstruction interpretation. Active $n$ is the number of clips retained for the same-camera reconstruction audit. MedGS LPIPS, CEMR, SF gap, and deformation energy are reported as medians over active clips only, preventing under-acquired near-static videos from being rewarded with low static-rendering error simply because they provide little acquisition evidence. SF gap denotes the static-vs-flexible LPIPS gap $\Delta_{\mathrm{stat}}=E^{\mathrm{LPIPS}}_{\mathrm{med}}-E^{\mathrm{LPIPS}}_{\mathrm{fit}}$. The corresponding unfiltered full-set profile, including raw Flow and Turn descriptors and metrics over all clips, is reported in Supplementary Table~A3.}
\label{tab:continuous-profile}
\label{tab:reconstruction-profile-summary}
\end{table}

\subsection{\texorpdfstring{Protocols}{Protocols}}
\label{sec:protocols}

The main reported protocol is a fair-format observed-frame audit over 3840 completed reconstructions.
It evaluates reconstruction quality along the camera path estimated from each generated clip, because the intended downstream use case is direct Gaussian reconstruction from generated video.
All clips are standardized to 49 frames, $256\times256$, and 8 fps before entering the fixed VGGT+DeformableGS+MedianGS stack.
We additionally retain two sanity protocols.
A seed-0 native-format run checks whether the fair-format profile is dominated by format conversion.
A 52-clip interleaved held-out pilot withholds interleaved frames during Gaussian optimization and measures MedianGS rendering error on the omitted frames, checking whether the all-frame profile changes when evaluated frames are not all used for fitting.
These checks accompany the main fair-format audit as protocol sensitivity evidence.

\begin{figure}[t]
\centering
\includegraphics[width=1.0\linewidth]{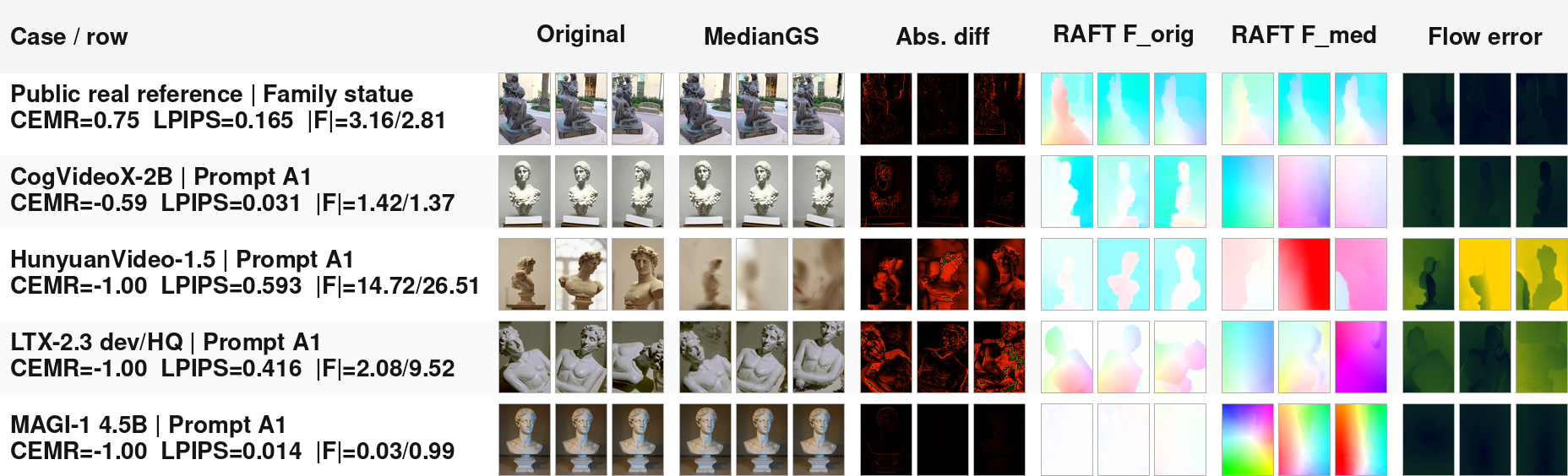}
\caption{CEMR visualization on a public-real reference and Prompt A1 examples. Rows compare input frames, MedianGS renders, absolute differences, input/render RAFT flow, and flow-error heatmaps under a shared color scale. The real reference has similar input/render flow. CogVideoX-2B and MAGI-1 have low MedianGS LPIPS but mismatched render flow, while HunyuanVideo-1.5 and LTX-2.3 HQ show both appearance and flow errors; CEMR therefore exposes temporal-motion mismatch that per-frame LPIPS can hide.}
\label{fig:cemr-a1-main-cases}
\end{figure}

\section{Experiments}
\label{sec:experiments}

\subsection{Setup}
\label{sec:exp-setup}

We evaluate fair-format reconstructions for 3840 generated clips over the 80 \gecoeval static-scene prompts, 12 open-weight model configurations, and four generation seeds $\{0,1,2,3\}$.
For each clip, we compute video-evidence descriptors, official \geco local geometry scores~\citep{gu2025geco}, VGGT camera descriptors~\citep{wang2025vggt}, DeformableGS rendering errors~\citep{yang2024deformable}, MedianGS static-reconstruction errors, deformation energy, and diagnostic visualizations.
The tables report the continuous fair-format reconstruction profile.
Seed-0 native-format and held-out results are retained as protocol checks.
The experiment section is organized by purpose: first the model-level reconstruction profile, then diagnostic disagreements within that profile, followed by protocol/backend sanity checks and ControlBench stress tests.

\subsection{Model-Level Reconstruction Profile}
\label{sec:exp-reconstruction-profile}

Table~\ref{tab:continuous-profile} is the main result table.
Each row pools four generation seeds and 80 prompts for one model.
The table is a staged audit rather than a sample-dropping filter: under-acquired clips are counted explicitly before reconstruction interpretation, and the reconstruction columns summarize only active clips so that near-static clips are not rewarded with low static-rendering error merely because they provide little acquisition evidence.
The unfiltered full-set profile, including raw Flow and Turn descriptors and metrics over all clips, is reported in Supplementary Table~A3.

Table~\ref{tab:continuous-profile} shows why one reconstruction score is misleading.
MAGI-1 has low MedianGS LPIPS but the largest under-acquisition rate and poor active-clip CEMR; HunyuanVideo-1.5 is often active but has much higher MedianGS LPIPS and deformation energy; and LTX-Video/SkyReels have low static errors but negative CEMR.
We therefore read the table as a split profile rather than a scalar ranking.

\subsection{Diagnostic Disagreement and GeCo Comparison}
\label{sec:diagnostic-breakdown}

\begin{figure}[t]
\centering
\includegraphics[width=0.8\linewidth]{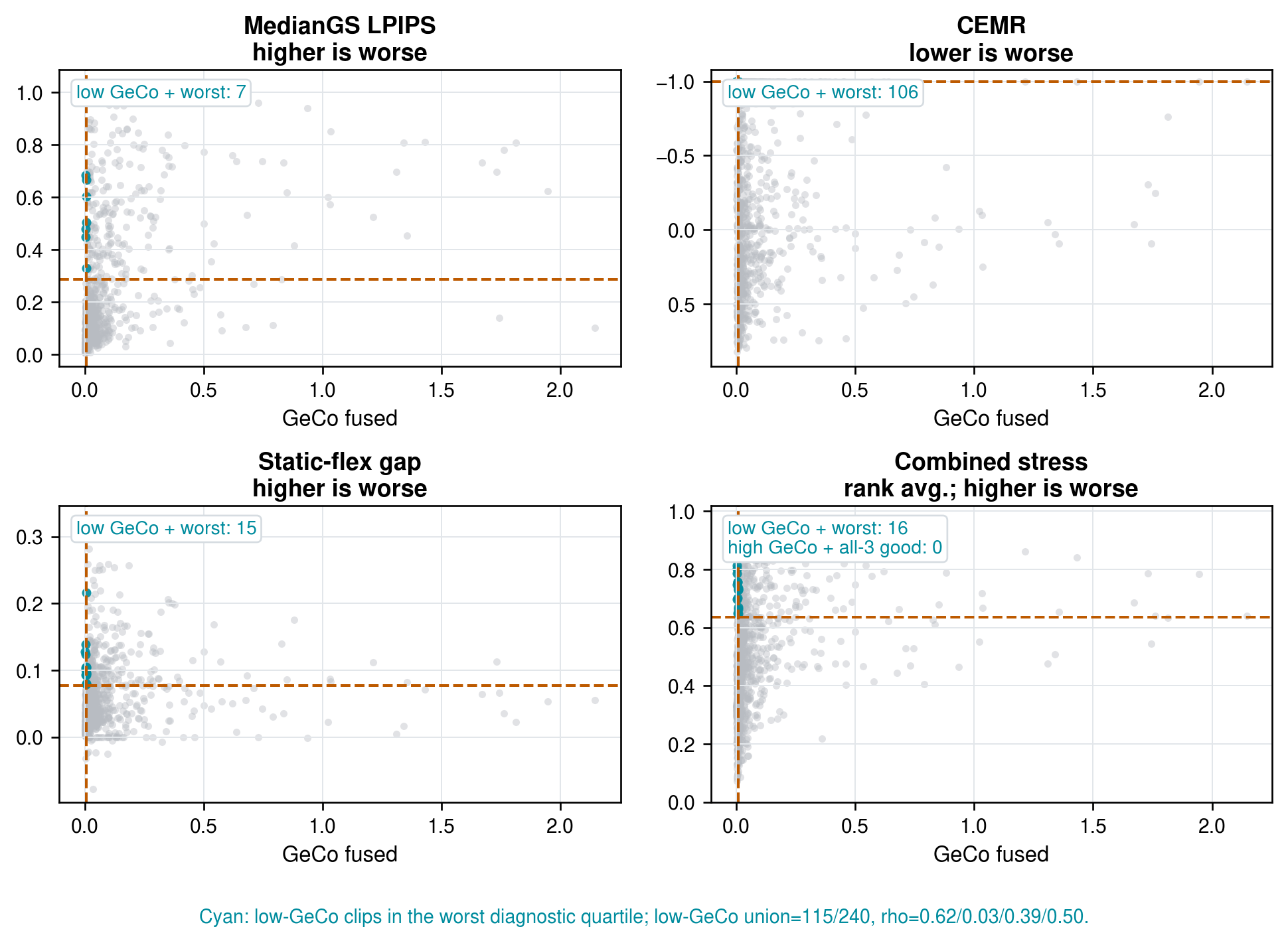}
\caption{\geco local-geometry scores and reconstruction diagnostics are complementary on the 960-clip seed-0 overlap. Cyan points have low \geco fused inconsistency but worst-quartile reconstruction stress. The right panel rank-averages MedianGS LPIPS, $-\mathrm{CEMR}$, and static-flex gap, showing that \geco is useful local context but not a replacement for the reconstruction audit.}
\label{fig:geco-vs-reconstruction-diagnostics}
\end{figure}

The diagnostics are non-interchangeable: MedianGS LPIPS measures per-frame static rendering, CEMR measures temporal-flow agreement, and the static-flex gap measures whether the flexible fit is much easier than the static proxy.
Figure~\ref{fig:cemr-a1-main-cases} shows that low LPIPS can still have mismatched render flow, so CEMR separates frame appearance from temporal-motion explanation.
We use \geco as an external local-geometry comparison.
On the seed-0 \geco-overlap subset, low \geco fused inconsistency can still coincide with reconstruction stress, especially CEMR-dominated failures; no high-\geco-inconsistency clip is simultaneously in the best quartile of MedianGS LPIPS, CEMR, and static-flex gap.
Figure~\ref{fig:geco-vs-reconstruction-diagnostics} and Supplementary Table~A12 report the correlations~\citep{spearman1961proof} and quartile counts; Supplementary Table~A15 gives the non-exclusive diagnostic breakdown.

\subsection{Protocol and Backend Sanity Checks}
\label{sec:protocol-checks}

Because the audit depends on VGGT-style camera estimation, Gaussian optimization, and optical flow, reconstruction residuals are backend-conditioned acquisition stress.
Supplementary checks cover camera geometry, static-proxy design, and flow backends: mean-collapse preserves MedianGS ordering on the 100-clip backend-bias subset ($\rho=0.997$), COLMAP/BA often registers high-stress clips, and Farneb{\"a}ck/GMFlow test CEMR sensitivity.
The fair-format audit is complete for 3840/3840 clips; native-format and held-out checks show useful but bounded stability, so we treat fair-format results as standardized acquisition stress rather than native-resolution video-quality rankings.

\subsection{\texorpdfstring{ControlBench as Metric Stress Test}{ControlBench as Metric Stress Test}}
\label{sec:controlbench}

ControlBench probes diagnostic semantics with 1000 public-real static sub-trajectories and 500 real-derived controls: frozen, digital zoom, lateral warp, texture repaint, and thin-plate spline foreground warp~\citep{bookstein2002principal}.
Supplementary Table~A4 and Figures~A8--A9 show that frozen clips collapse motion/CEMR, lateral warp and texture repaint raise MedianGS LPIPS, foreground warp increases the static-flex gap, and digital zoom remains borderline.

\section{Discussion and Limitations}
\label{sec:discussion}

\paragraph{Scope.}
\bench reports reconstruction diagnostics, not visual-quality rankings or ground-truth labels; low/high error can reflect acquisition evidence, motion, inconsistency, appearance, occlusion, or backend stress.
We will release code, manifests, metrics, diagnostic flags, ControlBench metadata, and license-safe indices/hashes.

\paragraph{Backend and controls.}
The audit depends on VGGT-style camera estimation and Gaussian optimization.
MASt3R, COLMAP/BA, mean-collapse, direct static 3DGS, GMFlow, and ControlBench test this sensitivity; MedianGS is a stable same-camera static proxy paired with the flexible fit, not a uniquely optimal aggregation.
The Prompt-A1 direct static-3DGS panel is qualitative only and does not claim uniform MedianGS superiority.

\paragraph{Metric limitations.}
CEMR is scalable but compact: it uses RAFT-Large on three sampled adjacent-frame pairs and mean residual magnitude, so local direction errors or long drift can be missed.
Deformation energy uses only position offsets, and MedianGS follows DeformableGS additive quaternion offsets with component-wise aggregation and renormalization rather than an intrinsic SO(3) median.
Denser flow checks, richer deformation terms, and rotation-aware aggregation are natural extensions.

\section{Conclusion}
\label{sec:conclusion}

Across 3840 reconstructions, \bench shows that single-score evaluation is misleading: low static error may reflect weak acquisition evidence, active clips can stress static reconstruction, and frame-level fit can miss temporal motion.
We therefore report a reconstruction-evidence profile rather than a one-residual ranking.

\FloatBarrier
\appendix
\setcounter{figure}{0}
\setcounter{table}{0}
\renewcommand{\thefigure}{\thesection\arabic{figure}}
\renewcommand{\thetable}{\thesection\arabic{table}}
\renewcommand{\theHfigure}{\thesection.\arabic{figure}}
\renewcommand{\theHtable}{\thesection.\arabic{table}}

\section{Appendix: Compact Evidence}
\label{app:additional}

This supplementary appendix keeps compact evidence needed to interpret the continuous reconstruction profile: the generation protocol, prompt-category metric breakdowns, ControlBench stress-test behavior, trajectory diagnostics, backend sanity checks, and representative reconstruction diagnostics.

\paragraph{Reconstruction-profile definition.}
For each generated clip, \bench reports active-acquisition descriptors, MedianGS static rendering error, static-render flow agreement, the static-vs-flexible LPIPS gap, and deformation energy.
CEMR is computed with RAFT-Large on uniformly sampled adjacent frame pairs at the fair-format $256\times256$ resolution and interpreted jointly with apparent flow as a static-render flow-agreement diagnostic.
\geco is reported later as an external local-geometry comparison, specifically to test whether local-geometry scores miss reconstruction failures captured by MedianGS LPIPS, CEMR, or the static-vs-flexible gap.

\paragraph{Generation panel and seed protocol.}
The fair-format model panel uses the 80 \gecoeval static-scene camera prompts, 12 open-weight model configurations, and four generation seeds: 0, 1, 2, and 3.
Each model therefore contributes $80\times4=320$ clips, for 3840 clips total.
The generated videos are first produced in each model's native generation setting and recorded in the release manifests with model metadata and hashes.
For the reconstruction audit, every clip is normalized to a common fair format: 49 frames, 8 fps, and $256\times256$ resolution, using the same temporal sampling and spatial resize/crop policy before VGGT, DeformableGS, MedianGS, CEMR, and diagnostic aggregation.
The native seed-0 audit gives a protocol-sensitivity check against this fair-format profile.

\begin{table}[!htbp]
\centering
\scriptsize
\resizebox{\linewidth}{!}{%
\begin{tabular}{ll}
\toprule
Item & Setting \\
\midrule
Prompt set & 80 controlled static-scene \gecoeval camera prompts \\
Generation seeds & 0, 1, 2, 3 \\
Model panel & 12 open-weight text-to-video configurations \\
Clips per model & 320 = 80 prompts $\times$ 4 seeds \\
Total fair-format clips & 3840 \\
Fair-format audit input & 49 frames, 8 fps, $256\times256$ resolution \\
Generation metadata & Native video settings, model metadata, and hashes stored in release manifests \\
\bottomrule
\end{tabular}}
\caption{Generation and seed protocol for the main fair-format audit. The table describes the model-panel clips included in the continuous reconstruction profile and separates them from ControlBench and held-out protocol checks.}
\label{tab:generation-seed-protocol}
\end{table}

\paragraph{Additional diagnostic tables.}
The appendix is organized by diagnostic role.
Tables~\ref{tab:prompt-category-breakdown}--\ref{tab:controlbench-stress-profile} summarize prompt-category behavior, the unfiltered model-level full-set profile, and ControlBench metric stress tests.
Tables~\ref{tab:vggt-backend-diagnostics}--\ref{tab:camera-family-diagnostics} report VGGT coverage and prompt-family trajectory diagnostics.
Tables~\ref{tab:mast3r-backend-ablation}--\ref{tab:flow-backend-ablation} and Figures~\ref{fig:colmap-ba-ablation}--\ref{fig:gmflow-cemr-ablation} are the backend-sanity block: MASt3R, COLMAP/BA, MedianGS-vs-mean collapse, image-space static-proxy negative controls, the Prompt-A1 direct static-3DGS baseline, and RAFT/Farneb{\"a}ck/GMFlow CEMR sensitivity.
Tables~\ref{tab:metric-disagreement}--\ref{tab:failure-breakdown} then report metric relationships, \geco complementarity, and non-exclusive failure decomposition; Table~\ref{tab:protocol-checks} gives protocol sanity checks.

\begin{table}[!htbp]
\centering
\scriptsize
\resizebox{1.0\linewidth}{!}{%
\begin{tabular}{lr*{5}{r}}
\toprule
Prompt group & Clips & Flow $\uparrow$ & Turn $\downarrow$ & MedianGS LPIPS $\downarrow$ & CEMR $\uparrow$ & Static-flex gap $\downarrow$ \\
\midrule
Object-centric & 960 & 5.15 & 164 & 0.098 & -0.66 & 0.042 \\
Indoor & 960 & 5.78 & 60 & 0.116 & -0.01 & 0.044 \\
Outdoor & 960 & 5.50 & 46 & 0.111 & 0.06 & 0.036 \\
Slow prompts & 1920 & 5.25 & 64 & 0.086 & -0.04 & 0.034 \\
Fast prompts & 1920 & 5.85 & 169 & 0.205 & -0.26 & 0.053 \\
\bottomrule
\end{tabular}}
\caption{Prompt-category continuous profile on the fair-format four-seed audit. Values are medians over clips in each group. The table is descriptive: it reports how acquisition style changes apparent flow, smoothed trajectory turn, MedianGS rendering error, CEMR, and static-vs-flexible gap without binarizing the clips.}
\label{tab:prompt-category-breakdown}
\end{table}

\begin{table}[!htbp]
\centering
\scriptsize
\resizebox{\linewidth}{!}{%
\begin{tabular}{lr*{6}{r}}
\toprule
Model & $n$ & Flow $\uparrow$ & Turn $\downarrow$ & MedianGS LPIPS $\downarrow$ & CEMR $\uparrow$ & Static-flex gap $\downarrow$ & Def. energy \\
\midrule
CogVideoX-2B & 320 & 6.21 & 53 & 0.101 & 0.12 & 0.039 & 0.107 \\
CogVideoX-5B & 320 & 5.96 & 55 & 0.120 & 0.06 & 0.047 & 0.115 \\
CogVideoX1.5-5B & 320 & 3.85 & 59 & 0.088 & -0.30 & 0.038 & 0.097 \\
HunyuanVideo & 320 & 3.18 & 204 & 0.118 & -1.00 & 0.040 & 0.127 \\
HunyuanVideo-1.5 & 320 & 7.15 & 121 & 0.351 & 0.06 & 0.060 & 0.211 \\
LTX-2.3 dev/HQ & 320 & 7.16 & 144 & 0.224 & 0.10 & 0.066 & 0.155 \\
LTX-2.3 dev/default & 320 & 6.02 & 205 & 0.298 & -0.01 & 0.049 & 0.167 \\
LTX-Video 13B & 320 & 6.19 & 71 & 0.075 & -0.15 & 0.035 & 0.084 \\
MAGI-1 4.5B & 320 & 1.60 & 214 & 0.035 & -1.00 & 0.018 & 0.076 \\
Open-Sora 2.0 & 320 & 4.02 & 100 & 0.117 & -0.08 & 0.044 & 0.101 \\
SkyReels-V2 & 320 & 4.52 & 49 & 0.066 & -0.24 & 0.027 & 0.094 \\
WAN 2.2 & 320 & 6.83 & 114 & 0.217 & 0.12 & 0.050 & 0.163 \\
\bottomrule
\end{tabular}}
\caption{Full-set continuous profile before active-evidence filtering. Values are medians over all 320 clips per model and include the raw apparent-flow and smoothed-trajectory-turn descriptors used before reconstruction interpretation. Main-paper Table 1 reports the staged view: under-acquisition is counted explicitly, while reconstruction diagnostics are summarized over active clips only.}
\label{tab:full-set-continuous-profile}
\end{table}

\begin{table}[!htbp]
\centering
\scriptsize
\resizebox{1.0\linewidth}{!}{%
\begin{tabular}{lrrrrr}
\toprule
Control family & $n$ & Flow $\uparrow$ & MedianGS LPIPS $\downarrow$ & CEMR $\uparrow$ & Static-flex gap $\downarrow$ \\
\midrule
Public real & 1000 & 2.66 & 0.093 & 0.15 & 0.013 \\
Digital zoom & 100 & 3.38 & 0.098 & 0.47 & 0.015 \\
Frozen & 100 & 0.00 & 0.029 & -1.00 & -0.000 \\
Lateral warp & 100 & 5.64 & 0.164 & 0.31 & 0.019 \\
Texture repaint & 100 & 2.63 & 0.199 & 0.10 & 0.028 \\
TPS foreground warp & 100 & 2.68 & 0.100 & 0.23 & 0.024 \\
\bottomrule
\end{tabular}}
\caption{ControlBench stress-test profile over the 1000 public-real / 500 transformed-control suite. Values are medians within each family. The public-real rows provide real camera-acquisition references; frozen controls suppress apparent flow, digital zoom and lateral warp stress motion/reconstruction agreement, texture repaint raises appearance error, and foreground warp probes deformation-like changes.}
\label{tab:controlbench-stress-profile}
\end{table}

\begin{table}[!htbp]
\centering
\scriptsize
\setlength{\tabcolsep}{4pt}
\begin{tabular}{p{0.26\linewidth}p{0.35\linewidth}p{0.31\linewidth}}
\toprule
Quantity & Setting / threshold & Role \\
\midrule
Apparent flow $\bar F_{\mathrm{orig}}$ & $\ge 0.2604628$ px/frame; control-driven weak-no-acquisition split retains 500/500 public-real positives and rejects 500/500 weak controls & Active-evidence gate; rejects visually near-static clips \\
Cumulative turn $\Theta_s$ & $\le 360.0^\circ$ after 10-point arc-length resampling of the smoothed trajectory & Active-evidence gate; rejects irregular multi-turn low-frequency trajectories \\
Boolean active rule & Flow condition AND cumulative-turn condition; otherwise the clip is counted as under-acquired before reconstruction interpretation & Defines which clips enter the active-only reconstruction summaries in main-paper Table 1 \\
Smoothed path length $\ell_s/(T-1)$ & No threshold; reported as context only & Non-degenerate-path diagnostic, not a gate term \\
Raw-to-smoothed jitter $\rho_j$ & No threshold; reported as context only & Camera-instability diagnostic; not used for rejection because VGGT center jitter can be coupled with FOV changes \\
FOV variation & No threshold; reported as context only & Camera-intrinsics instability diagnostic \\
Trajectory centering / normalization & Center by mean camera center; visualization JSON is divided by $s=\max(\mathrm{P}_{95}$ radius, max coordinate range, $\epsilon)$ & Descriptor-only scale normalization; not written back to camera files \\
Savitzky--Golay window & Largest odd window up to 11 frames, at least 3 frames & Descriptor-only local smoothing \\
Savitzky--Golay polynomial order & $\min(3,\mathrm{window}-1)$ with interpolation mode & Descriptor-only local smoothing \\
Arc-length samples & 10 equally spaced samples on the smoothed curve & Computes cumulative-turn descriptor \\
\bottomrule
\end{tabular}
\caption{Active-evidence gate and trajectory-descriptor settings. The active gate uses apparent motion and low-frequency trajectory direction coherence only; smoothed path length, raw-to-smoothed jitter, and FOV variation are reported as context. The smoothed trajectory is used only for active-evidence descriptors and visualization; DeformableGS and MedianGS use the original VGGT-estimated intrinsics and poses.}
\label{tab:active-gate-settings}
\end{table}

\begin{table}[!htbp]
\centering
\scriptsize
\setlength{\tabcolsep}{4pt}
\begin{tabular}{p{0.27\linewidth}p{0.66\linewidth}}
\toprule
Component & Fixed setting \\
\midrule
Fair-format input & 49 frames at 8 fps, resized so the shorter side is 256 px and then center-cropped to $256\times256$. \\
VGGT input & All 49 fair-format frames; $518\times518$ VGGT inference; depth confidence threshold 3; at most 120k depth-derived COLMAP points. \\
Camera/geometry initialization & VGGT-estimated intrinsics and poses exported as a 49-image COLMAP text model and initialized with the VGGT point cloud. \\
DeformableGS training & 10,000 iterations; COLMAP-format image/camera input; spherical-harmonic degree 3; black background; random train-camera sampling. \\
Optimizer and learning rates & Adam~\citep{kingma2014adam} with $\epsilon=10^{-15}$ for Gaussian and deformation parameters. The schedule uses position lr init/final $1.6{\times}10^{-4}/1.6{\times}10^{-6}$ with delay multiplier 0.01; the implementation applies spatial scale 5 to Gaussian xyz and to the initial deformation lr. Feature-DC, feature-rest, opacity, scale, and rotation learning rates are $2.5{\times}10^{-3}$, $1.25{\times}10^{-4}$, $5{\times}10^{-2}$, $10^{-3}$, and $10^{-3}$. \\
Training loss & $(1-\lambda_{\mathrm{DSSIM}})L_1+\lambda_{\mathrm{DSSIM}}(1-\mathrm{SSIM})$ with $\lambda_{\mathrm{DSSIM}}=0.2$ and SSIM from~\citet{wang2004image}. \\
Deformation warm-up & Deformation offsets are zero for the first 3,000 iterations; the learned deformation field is used afterwards. \\
Densification / pruning & Densification statistics are collected while iteration $<15{,}000$; densify/prune every 100 iterations after iteration 500 with gradient threshold $7{\times}10^{-4}$ and percent-dense 0.01; opacity reset every 3,000 iterations. Since fair runs stop at 10,000 iterations, this policy remains active through training. \\
DeformableGS render & Latest saved iteration rendered on the train cameras; fair-format static metrics use the train split for the dynamic-COLMAP outputs. \\
MedianGS proxy & Temporal median of full DeformableGS deltas over all 49 native fair-format frames; rendered at iteration 10,000 along the original VGGT-estimated intrinsics and poses. \\
Evaluation frames $\mathcal{S}$ & All 49 matched fair-format frames for LPIPS, static-vs-flexible gap, PSNR, SSIM, and static render-flow context; LPIPS uses the AlexNet backbone from~\citet{zhang2018unreasonable} at native fair resolution. \\
Deformation-energy context & 16 uniformly spaced normalized time samples, at most 50k Gaussians, chunk size 32768. \\
CEMR & Torchvision RAFT-Large at $256\times256$ on three uniformly sampled adjacent-frame pairs. \\
Runtime reporting & Full per-clip runtime was not aggregated as a benchmark field. All reported clips use the same fixed reconstruction settings. \\
\bottomrule
\end{tabular}
\caption{Implementation settings for the fair-format reconstruction profile. Values are fixed by the fair-format audit configuration and DeformableGS optimizer defaults; runtime is reported as context rather than as a benchmark metric.}
\label{tab:implementation-settings}
\end{table}

\paragraph{Trajectory descriptor details.}
Let $\mathbf{c}_t$ be the original VGGT camera center at frame $t$.
For descriptor computation only, we center the trajectory and robustly normalize it by
\[
s=\max\left(\mathrm{P}_{95}\left(\lVert \mathbf{c}_t-\bar{\mathbf{c}}\rVert_2\right),
\max_k \mathrm{range}(c_{t,k}), \epsilon\right).
\]
Here $s$ is the descriptor-only normalization scale, $\bar{\mathbf{c}}$ is the mean camera center, $\mathrm{P}_{95}$ is the 95th percentile over frames, $c_{t,k}$ is coordinate $k$ of $\mathbf{c}_t$, and $\epsilon$ prevents division by zero.
Each coordinate is then smoothed with a Savitzky--Golay filter~\citep{savitzky1964smoothing} using the largest odd window up to 11 frames and polynomial order at most three.
The trajectory diagnostic uses apparent optical flow and the cumulative tangent turn $\Theta_s$ computed after resampling the smoothed curve to 10 equally spaced arc-length points.
When the image content is frozen or nearly static, VGGT may fit small inconsistent camera-center changes instead of a coherent sweep, so the smoothed trajectory can accumulate an abnormally large turn angle.
We report the smoothed path length per step $\ell_s/(T-1)$, the raw-to-smoothed jitter ratio $\rho_j=\ell_{\mathrm{raw}}/(\ell_s+\epsilon)$, and horizontal-FOV range as trajectory context.
This distinction matters because generated clips can induce coupled VGGT center and FOV jitter; penalizing center jitter alone can reject camera estimates that still render coherently when the original VGGT centers and intrinsics are used together.
Missing, degenerate, or non-finite trajectories are reported as failed camera-diagnostic cases rather than silently interpreted.
These diagnostics are never written back to the camera files; all DeformableGS and MedianGS reconstructions use the original VGGT-estimated camera poses and per-frame intrinsics.

\begin{figure}[!htbp]
\centering
\includegraphics[width=0.985\linewidth]{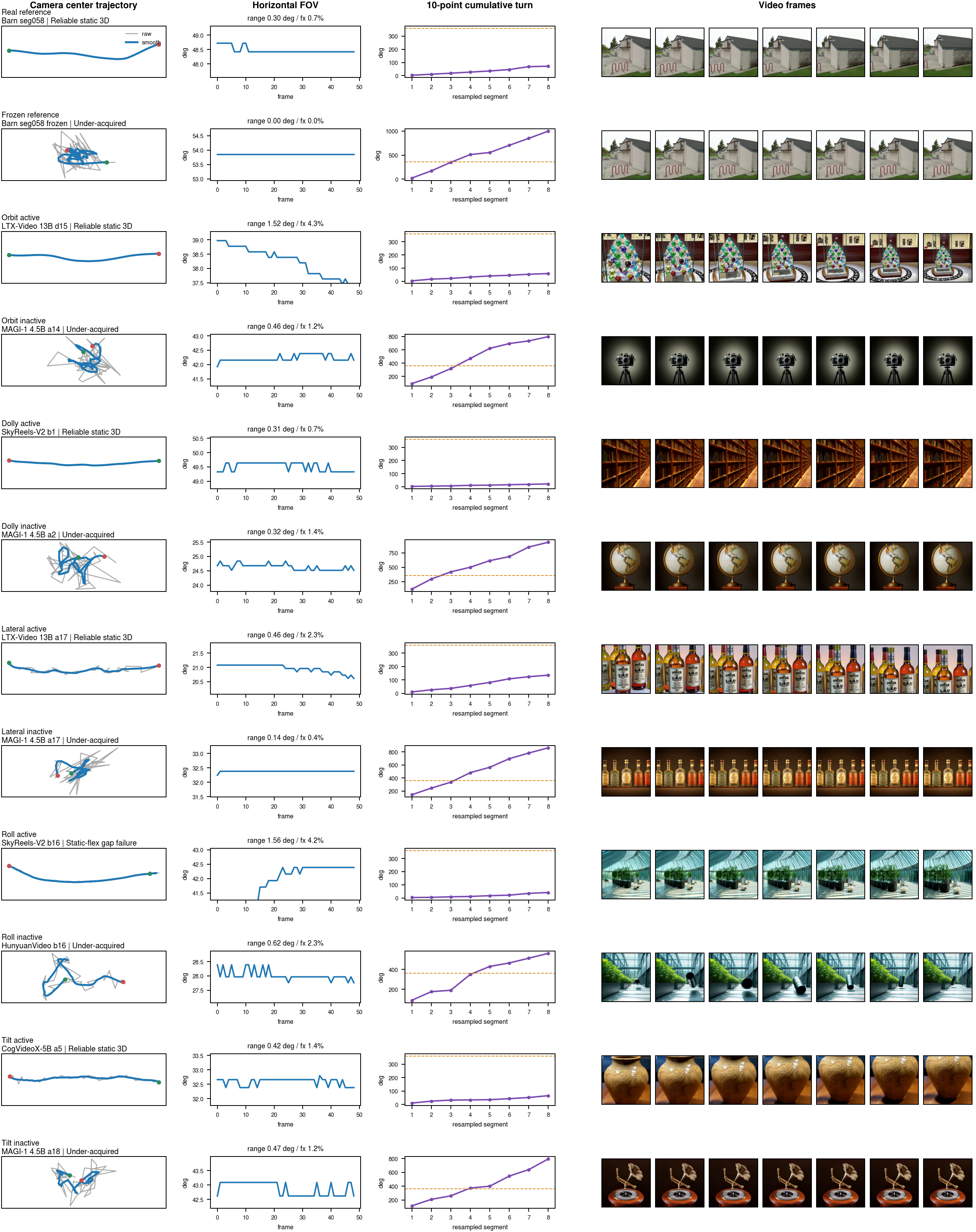}
\caption{VGGT camera diagnostics for a real reference/frozen-control pair and active/inactive generated examples from orbit, dolly/forward, lateral, roll, and tilt prompts. The columns show camera-center trajectories in PCA-relative coordinates (raw gray, smoothed blue, first/last green/red), per-frame horizontal FOV, 10-point cumulative turn with a one-revolution reference line, and seven uniformly sampled video frames. Each generated family includes one active-looking clip and one visually near-static clip. FOV panels share a common vertical span per row; smoothed trajectories are diagnostic only, and reconstruction uses the original VGGT-estimated poses and per-frame intrinsics.}
\label{fig:vggt-trajectory-diagnostics}
\end{figure}

\begin{table}[!htbp]
\centering
\scriptsize
\resizebox{0.6\linewidth}{!}{%
\begin{tabular}{l*{5}{r}}
\toprule
Scope & Clips & Seeds & Models & Pose ok & Metrics ok \\
\midrule
Seed-0 fair & 960 & 1 & 12 & 100.0 & 100.0 \\
Seed-1 fair & 960 & 1 & 12 & 100.0 & 100.0 \\
Seed-2 fair & 960 & 1 & 12 & 100.0 & 100.0 \\
Seed-3 fair & 960 & 1 & 12 & 100.0 & 100.0 \\
Combined fair & 3840 & 4 & 12 & 100.0 & 100.0 \\
\bottomrule
\end{tabular}}
\caption{Fair-format backend coverage for the current VGGT plus Gaussian audit stack. Pose status is 99.97\% exact-ok on the combined panel and rounds to 100.0\% at one decimal; static metrics are complete for all 3840 reconstructed clips. This is a coverage check, not an alternative-backend ablation.}
\label{tab:vggt-backend-diagnostics}
\end{table}

\begin{table}[!htbp]
\centering
\scriptsize
\resizebox{\linewidth}{!}{%
\begin{tabular}{lrrrrrrrrr}
\toprule
Family & Prompts & Clips & Active & Traj. comp. & Low-error among comp. & Low-error among non-comp. & Med. trans. & Med. rot. & Low motion \\
\midrule
dolly/forward & 30 & 1440 & 54.6 & 64.4 & 7.7 & 3.3 & 0.625 & 0.175 & 9.5 \\
lateral & 6 & 288 & 58.3 & 45.8 & 20.5 & 2.6 & 0.419 & 0.297 & 14.9 \\
mixed/other & 16 & 768 & 54.8 & 82.4 & 10.4 & 0.7 & 0.451 & 0.245 & 17.1 \\
orbit & 16 & 768 & 54.8 & 58.7 & 10.4 & 3.5 & 0.588 & 0.943 & 14.6 \\
roll & 1 & 48 & 60.4 & 45.8 & 13.6 & 0.0 & 0.675 & 0.757 & 8.3 \\
tilt & 11 & 528 & 56.4 & 64.4 & 10.9 & 4.8 & 0.401 & 0.688 & 13.6 \\
\bottomrule
\end{tabular}}
\caption{Minimal deterministic prompt-family trajectory compliance from VGGT-estimated trajectories. The rules summarize broad trajectory evidence--orbit rotation, dolly monotonicity, lateral dominance, roll angle, or tilt/vertical change--as descriptor-level context.}
\label{tab:camera-family-diagnostics}
\end{table}

The backend sanity checks below use descriptive diagnostic strata rather than method outputs: Low-static-error and Low-motion are score-based reference subsets, Low-stress denotes comparatively mild reconstruction diagnostics, Deformation-stress denotes cases selected for large flexible-vs-static behavior, and High-reconstruction-stress denotes active clips with high static reconstruction stress.

\begin{table}[!htbp]
\centering
\scriptsize
\resizebox{\linewidth}{!}{%
\begin{tabular}{lrrrrrr}
\toprule
Diagnostic stratum & $n$ & Pair ok & Pair geom. & Geom. pass & Geom. score & Med. flow \\
\midrule
Low-static-error subset & 5 & 100.0 & 100.0 & 100.0 & 0.77 & 0.149 \\
Low-motion subset & 53 & 100.0 & 64.2 & 54.7 & 0.51 & 0.051 \\
Low-stress subset & 18 & 100.0 & 100.0 & 94.4 & 0.55 & 0.132 \\
Deformation-stress subset & 14 & 100.0 & 78.6 & 78.6 & 0.38 & 0.199 \\
High-reconstruction-stress subset & 102 & 100.0 & 98.0 & 91.2 & 0.53 & 0.188 \\
\midrule
\multicolumn{7}{l}{Model-level rank agreement on the same subset.} \\
\multicolumn{2}{l}{Full model reconstruction profile vs. MASt3R geometry score} & \multicolumn{2}{r}{Spearman $\rho$=0.87} & \multicolumn{3}{r}{$n_\mathrm{models}=12$} \\
\multicolumn{2}{l}{Full low-static-error yield vs. MASt3R geometry pass} & \multicolumn{2}{r}{Spearman $\rho$=0.66} & \multicolumn{3}{r}{$n_\mathrm{models}=12$} \\
\multicolumn{2}{l}{Subset VGGT+GS low-static-error grouping vs. MASt3R geometry score} & \multicolumn{2}{r}{Spearman $\rho$=0.59} & \multicolumn{3}{r}{$n_\mathrm{models}=12$} \\
\multicolumn{2}{l}{Subset VGGT+GS high-stress grouping vs. MASt3R geometry score} & \multicolumn{2}{r}{Spearman $\rho$=-0.48} & \multicolumn{3}{r}{$n_\mathrm{models}=12$} \\
\multicolumn{2}{l}{Full active coverage vs. MASt3R pair geometry} & \multicolumn{2}{r}{Spearman $\rho$=0.47} & \multicolumn{3}{r}{$n_\mathrm{models}=12$} \\
\bottomrule
\end{tabular}}
\caption{Independent MASt3R backend ablation on a 192-clip fair-format subset (16 prompt--seed keys $\times$ 12 models). MASt3R provides an independent feed-forward matching/geometry backend and reports pairwise geometry validity, non-degenerate geometry-pass rate, and model-level agreement with the VGGT+Gaussian reconstruction profile.}
\label{tab:mast3r-backend-ablation}
\end{table}

\begin{table}[!htbp]
\centering
\scriptsize
\resizebox{\linewidth}{!}{%
\begin{tabular}{lrrrrrrr}
\toprule
Diagnostic stratum & $n$ & COLMAP ok & Reg.$\ge75\%$ & Med. reg. & Reproj. & Pts3D & Med. LPIPS \\
\midrule
Low-static-error subset & 23 & 95.7 & 34.8 & 50.0 & 0.40 & 272 & 0.083 \\
Low-motion subset & 23 & 82.6 & 26.1 & 25.0 & 0.26 & 103 & 0.064 \\
Low-stress subset & 20 & 75.0 & 25.0 & 33.3 & 0.43 & 114 & 0.226 \\
Deformation-stress subset & 10 & 60.0 & 0.0 & 16.7 & 0.39 & 58 & 0.645 \\
High-reconstruction-stress subset & 24 & 91.7 & 41.7 & 45.8 & 0.32 & 188 & 0.123 \\
\bottomrule
\end{tabular}}
\caption{Classic SfM/BA sanity check on the 100-clip stratified backend-bias subset using COLMAP SIFT matching, incremental SfM, and bundle adjustment~\citep{schonberger2016structure}. Reg.$\ge75\%$ counts clips where at least 75\% of the 12 sampled frames register. High-reconstruction-stress samples often obtain a COLMAP model, so high reconstruction stress is not simply VGGT pose dropout; it still reflects the full static-reconstruction and CEMR behavior.}
\label{tab:colmap-ba-sanity}
\end{table}

\begin{figure}[!htbp]
\centering
\includegraphics[width=\linewidth]{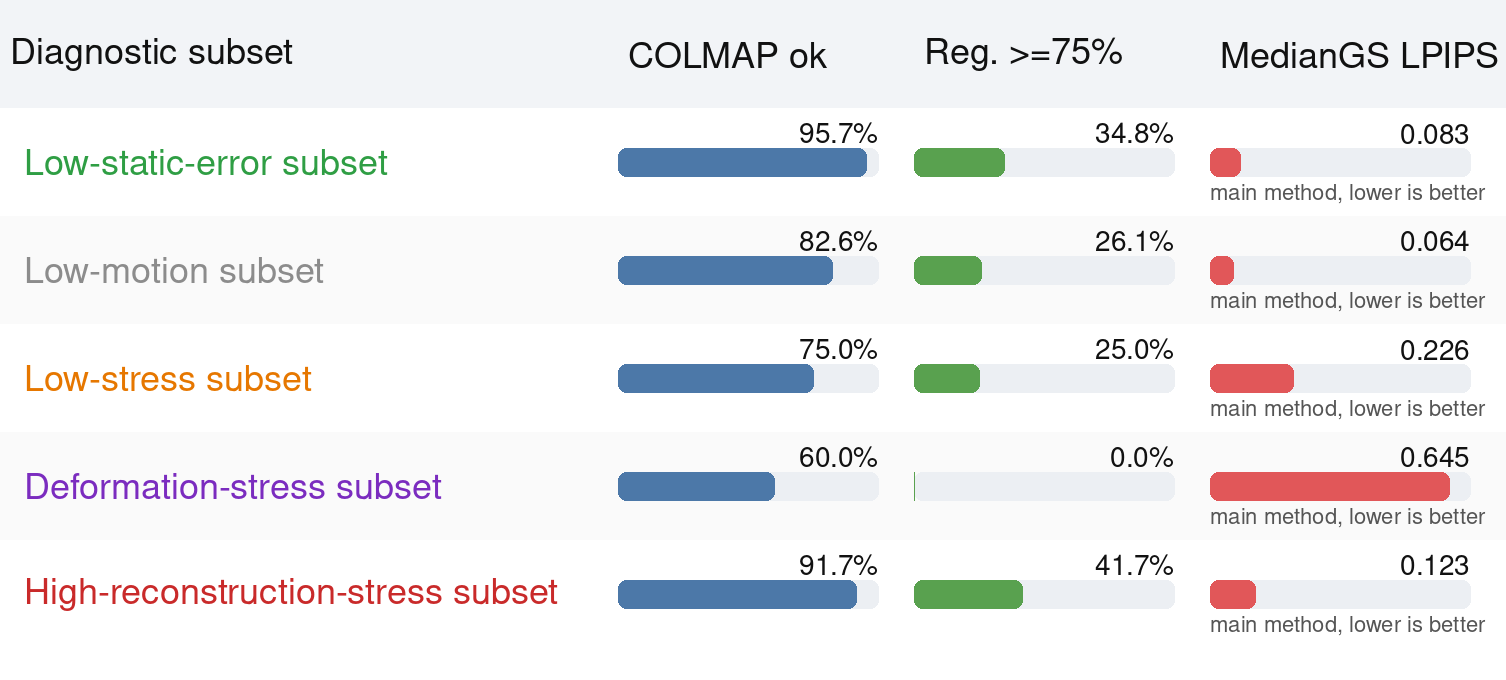}
\caption{COLMAP/BA ablation visualized against diagnostic sampling strata. Bars show whether a classic SfM/BA pipeline can still register sampled frames. Many high-stress generated clips still produce a COLMAP model, so their reconstruction difficulty is not explained by VGGT pose dropout alone; the continuous profile also inspects static-reconstruction and CEMR behavior.}
\label{fig:colmap-ba-ablation}
\end{figure}

\begin{table}[!htbp]
\centering
\scriptsize
\resizebox{0.78\linewidth}{!}{%
\begin{tabular}{lrrrr}
\toprule
Group & $n$ & MedianGS & Mean-collapse & $\Delta$ \\
\midrule
All subset & 100 & 0.119 & 0.121 & 0.001 \\
Low-static-error subset & 23 & 0.083 & 0.085 & -0.000 \\
Low-motion subset & 23 & 0.064 & 0.065 & 0.000 \\
Low-stress subset & 20 & 0.226 & 0.230 & 0.002 \\
Deformation-stress subset & 10 & 0.645 & 0.668 & 0.029 \\
High-reconstruction-stress subset & 24 & 0.123 & 0.124 & 0.001 \\
\midrule
MedianGS vs mean-collapse rank & 100 & \multicolumn{3}{r}{Spearman $\rho=0.997$} \\
\bottomrule
\end{tabular}}
\caption{Static-collapse sensitivity on the same 100-clip backend-bias subset. MedianGS uses temporal-median deformation deltas; mean-collapse replaces the temporal median by the temporal mean while keeping the VGGT-estimated intrinsics and poses, Gaussian attributes, and renderer fixed. Median and mean aggregation have nearly identical ranking and small median LPIPS shifts, so the static-reconstruction ordering is not a special artifact of choosing a temporal median rather than a temporal mean.}
\label{tab:static-collapse-ablation}
\end{table}

\begin{figure}[!htbp]
\centering
\includegraphics[width=0.86\linewidth]{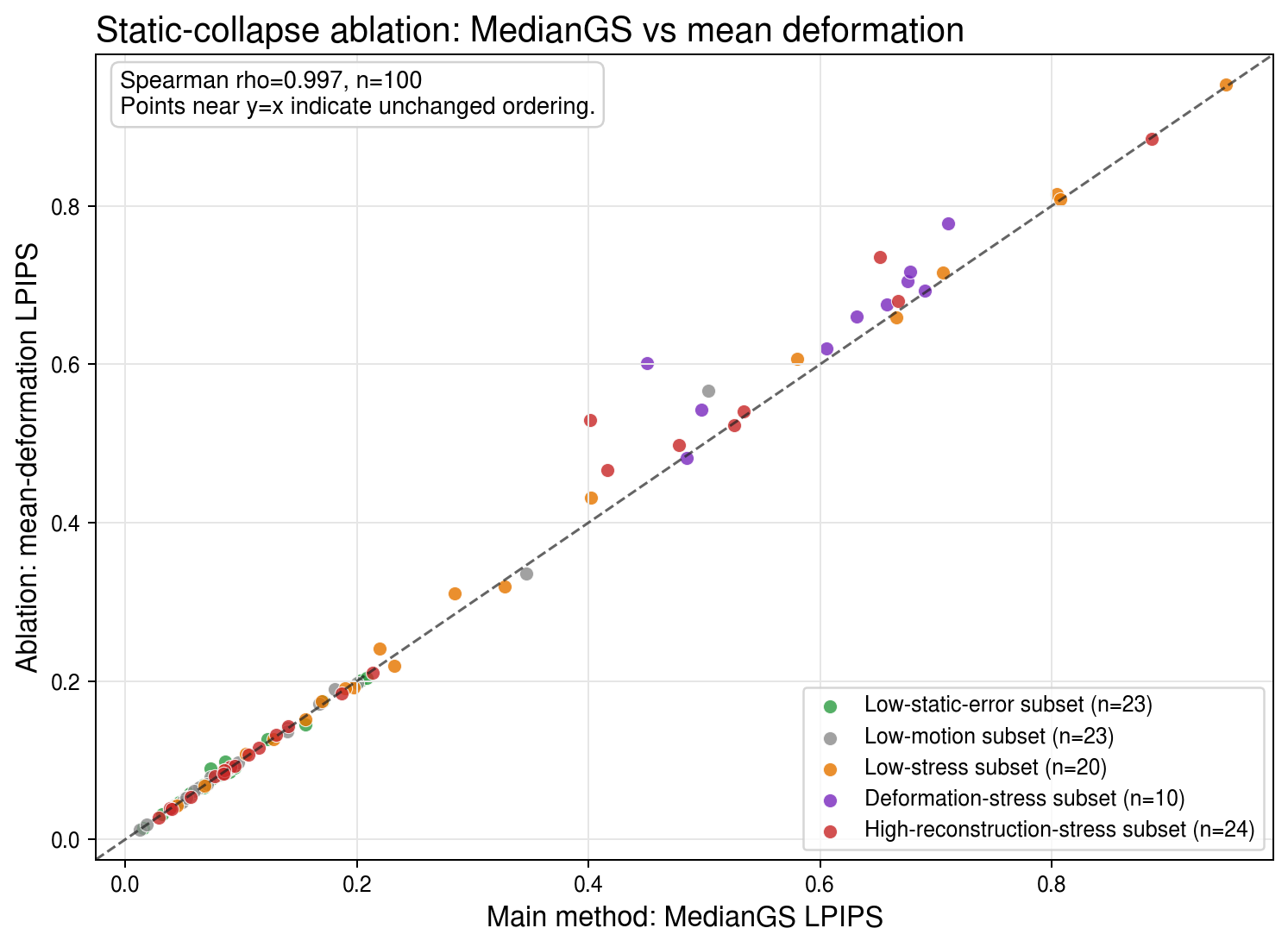}
\caption{Static-collapse ablation visualized sample by sample. The x-axis is the main MedianGS LPIPS residual and the y-axis replaces temporal-median deformation aggregation with temporal-mean aggregation. Points lie close to the diagonal and the rank agreement is high, showing that the main static-reconstruction ordering is stable to this aggregation-rule change.}
\label{fig:static-collapse-ablation}
\end{figure}

\begin{figure}[!htbp]
\centering
\includegraphics[width=0.90\linewidth]{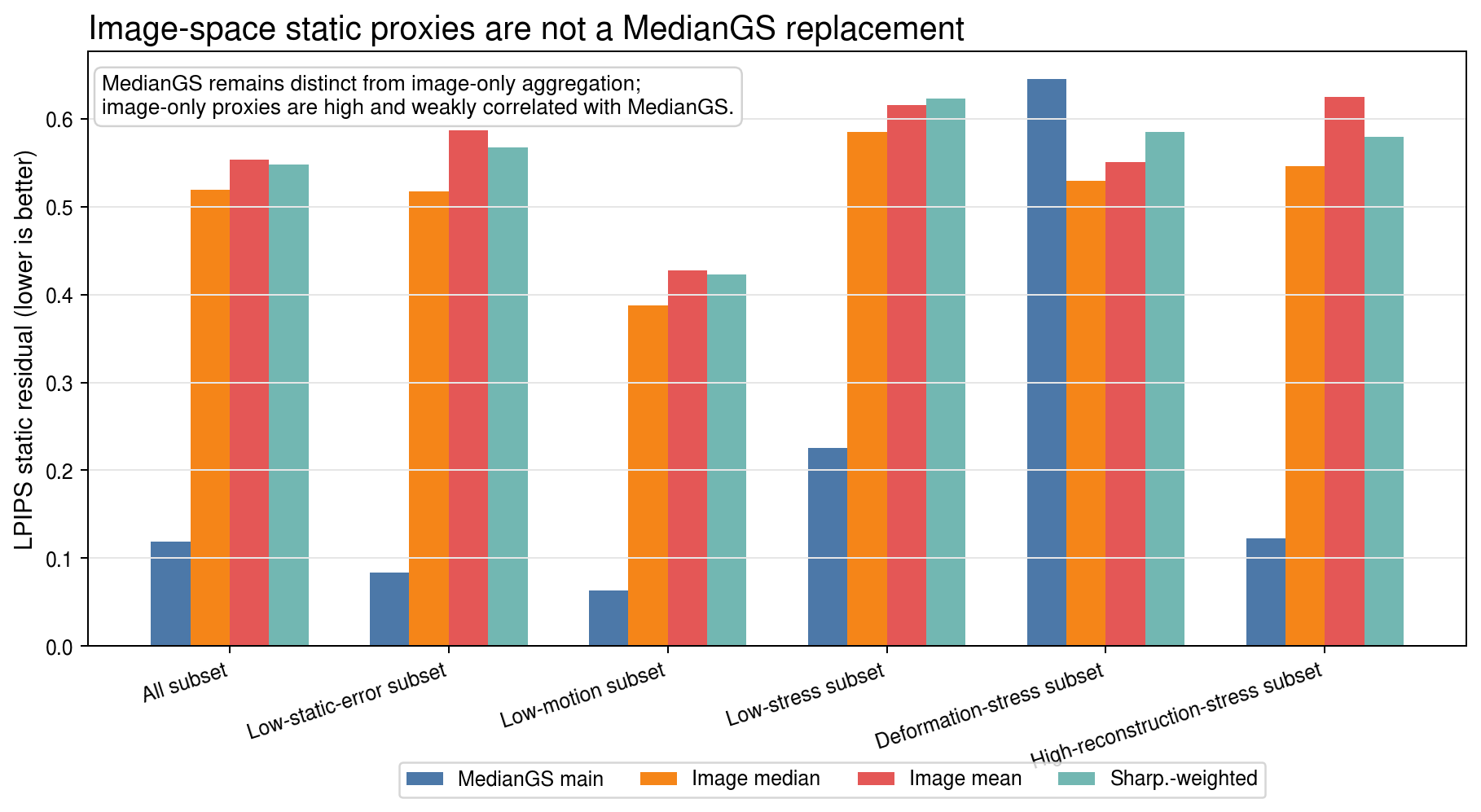}
\caption{Auxiliary image-space static proxy ablation. Image median, image mean, and sharpness-weighted image averages are much higher residual proxies and weakly correlated with MedianGS, showing that simple image aggregation lacks the same-camera Gaussian structure captured by MedianGS.}
\label{fig:image-static-proxy-ablation}
\end{figure}

\begin{figure}[!htbp]
\centering
\includegraphics[width=0.96\linewidth]{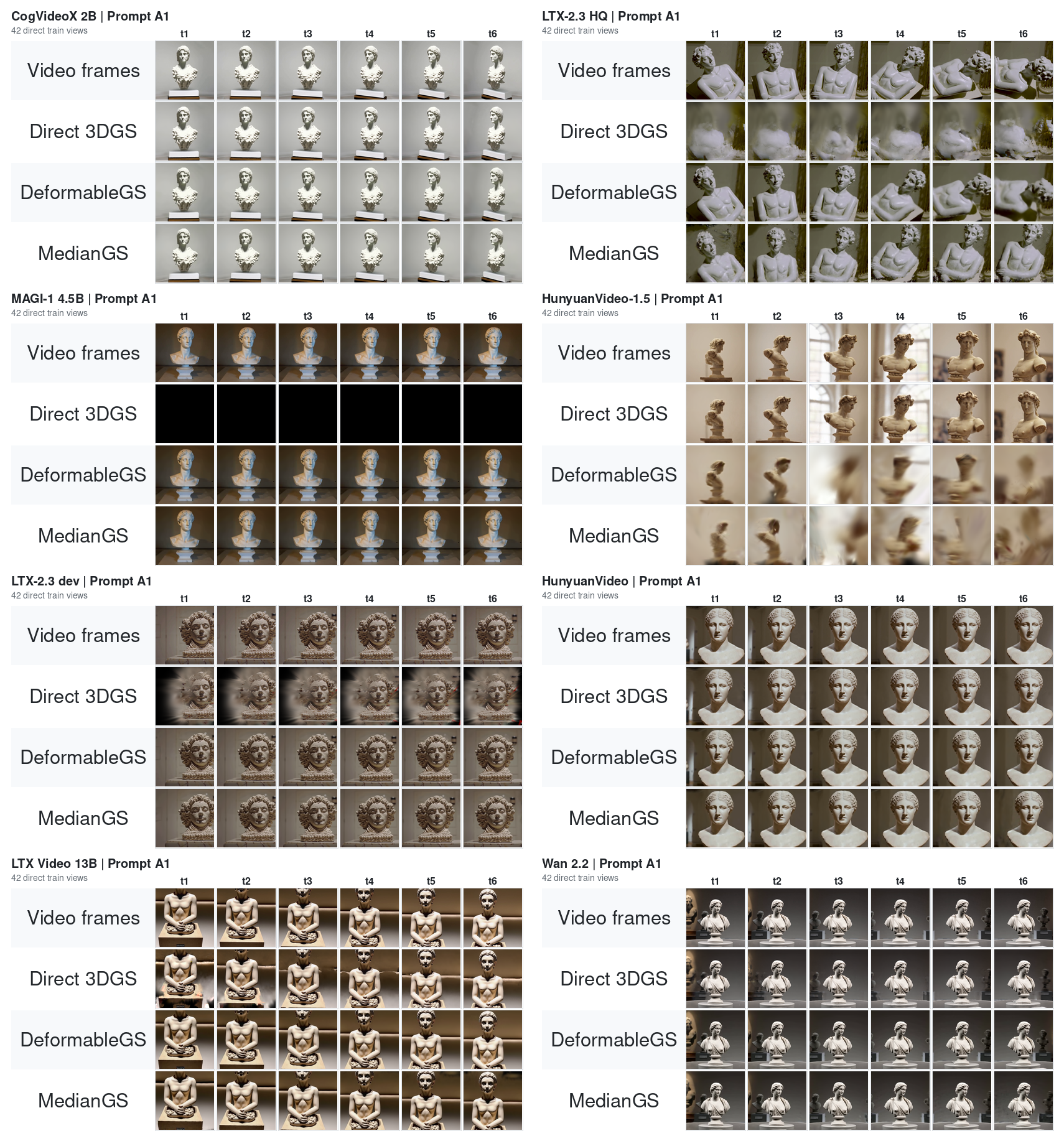}
\caption{Prompt A1 static-reconstruction baseline. Each block compares video frames, a vanilla static 3DGS trained directly from the same clip using the VGGT-estimated intrinsics and poses, the flexible DeformableGS render, and the MedianGS static proxy. Direct static 3DGS fails visibly for LTX-2.3 HQ, MAGI-1 4.5B, and LTX-2.3 dev, producing blurry blobs or black renders, while the DeformableGS and MedianGS rows remain usable. HunyuanVideo-1.5 shows the opposite stress case: direct static 3DGS appears clean, but the DeformableGS and MedianGS renderings blur. The other examples have smaller motion, more consistent structure, or cleaner video evidence, and all three reconstruction variants are visually closer to the video frames. Overall, the comparison supports using the DeformableGS-to-MedianGS reconstruction profile as a more robust diagnostic than direct static 3DGS reconstruction alone.}
\label{fig:direct-static3dgs-a1-baseline}
\end{figure}

\noindent Table~\ref{tab:metric-disagreement} uses Spearman's rank correlation coefficient $\rho$~\citep{spearman1961proof} to measure whether two quantities order clips or models similarly; it does not require the quantities to share units or a linear scale. ``Model Spearman'' is computed over the 12 model medians, while ``sample Spearman'' is computed over the 960 seed-0 clips with official \geco scores. The count rows use quartiles: low \geco fused inconsistency means the lowest \geco fused-score quartile, worst-quartile diagnostic means the worst 25\% for the displayed reconstruction diagnostic, combined reconstruction stress is the equal-weight rank average of MedianGS LPIPS, $-\mathrm{CEMR}$, and static-vs-flexible gap, and all-three-good means the best quartile for all three reconstruction diagnostics.

\begin{table}[!htbp]
\centering
\scriptsize
\resizebox{1.0\linewidth}{!}{%
\begin{tabular}{llrr}
\toprule
Comparison & Statistic & Value & $n$ \\
\midrule
Flow vs MedianGS LPIPS & model Spearman $\rho$ & 0.63 & 12 \\
Flow vs CEMR & model Spearman $\rho$ & 0.84 & 12 \\
MedianGS LPIPS vs CEMR & model Spearman $\rho$ & 0.59 & 12 \\
MedianGS LPIPS vs Static-flex gap & model Spearman $\rho$ & 0.96 & 12 \\
CEMR vs Static-flex gap & model Spearman $\rho$ & 0.69 & 12 \\
\midrule
\geco fused vs MedianGS LPIPS & sample Spearman $\rho$ & 0.62 & 960 \\
\geco fused vs CEMR & sample Spearman $\rho$ & 0.03 & 960 \\
\geco fused vs Static-flex gap & sample Spearman $\rho$ & 0.39 & 960 \\
\geco fused vs combined reconstruction stress & sample Spearman $\rho$ & 0.50 & 960 \\
Low \geco fused inconsistency with any worst-quartile diagnostic & count & 115/240 & 960 \\
Low \geco fused inconsistency with worst-quartile combined stress & count & 16/240 & 960 \\
High \geco fused inconsistency with all-three-good diagnostics & count & 0/240 & 960 \\
\bottomrule
\end{tabular}}
\caption{Metric relationships for the reconstruction profile. The first block uses model medians over the fair-format four-seed panel. The \geco block uses the 960-clip seed-0 overlap where official \geco scores are available and mirrors main-paper Figure 4. \geco has moderate agreement with MedianGS LPIPS, almost no sample-level agreement with CEMR, and partial agreement with the equal-weight combined reconstruction stress. The quartile counts summarize complementarity: clips with low \geco fused inconsistency can still occupy the worst reconstruction-diagnostic quartiles, while clips with high \geco fused inconsistency can have one good diagnostic but never land in the best quartile of all three reconstruction diagnostics simultaneously.}
\label{tab:metric-disagreement}
\end{table}

\begin{table}[!htbp]
\centering
\scriptsize
\resizebox{0.65\linewidth}{!}{%
\begin{tabular}{lrr}
\toprule
Comparison & $n$ & Value \\
\midrule
RAFT vs Farneb{\"a}ck CEMR Spearman & 3840 & 0.91 \\
RAFT vs Farneb{\"a}ck CEMR Pearson & 3840 & 0.90 \\
RAFT vs GMFlow CEMR Spearman & 100 & 0.79 \\
RAFT vs GMFlow CEMR Pearson & 100 & 0.80 \\
RAFT vs GMFlow sign-agreement around zero & 100 & 81.0 \\
GMFlow completion & 100 & 100.0 \\
\bottomrule
\end{tabular}}
\caption{Optical-flow-backend sensitivity for CEMR. The first block compares the reported RAFT-Large CEMR values against the earlier lightweight Farneb{\"a}ck variant~\citep{farneback2003two} on the full fair-format panel. The second block recomputes CEMR on the 100-clip backend-bias subset with GMFlow~\citep{xu2022gmflow}. The modern-flow cross-check gives moderate-to-high sample agreement, reducing the risk that CEMR is a RAFT-only artifact while showing that exact clip-level values remain backend-sensitive.}
\label{tab:flow-backend-ablation}
\end{table}

\begin{figure}[!htbp]
\centering
\includegraphics[width=0.86\linewidth]{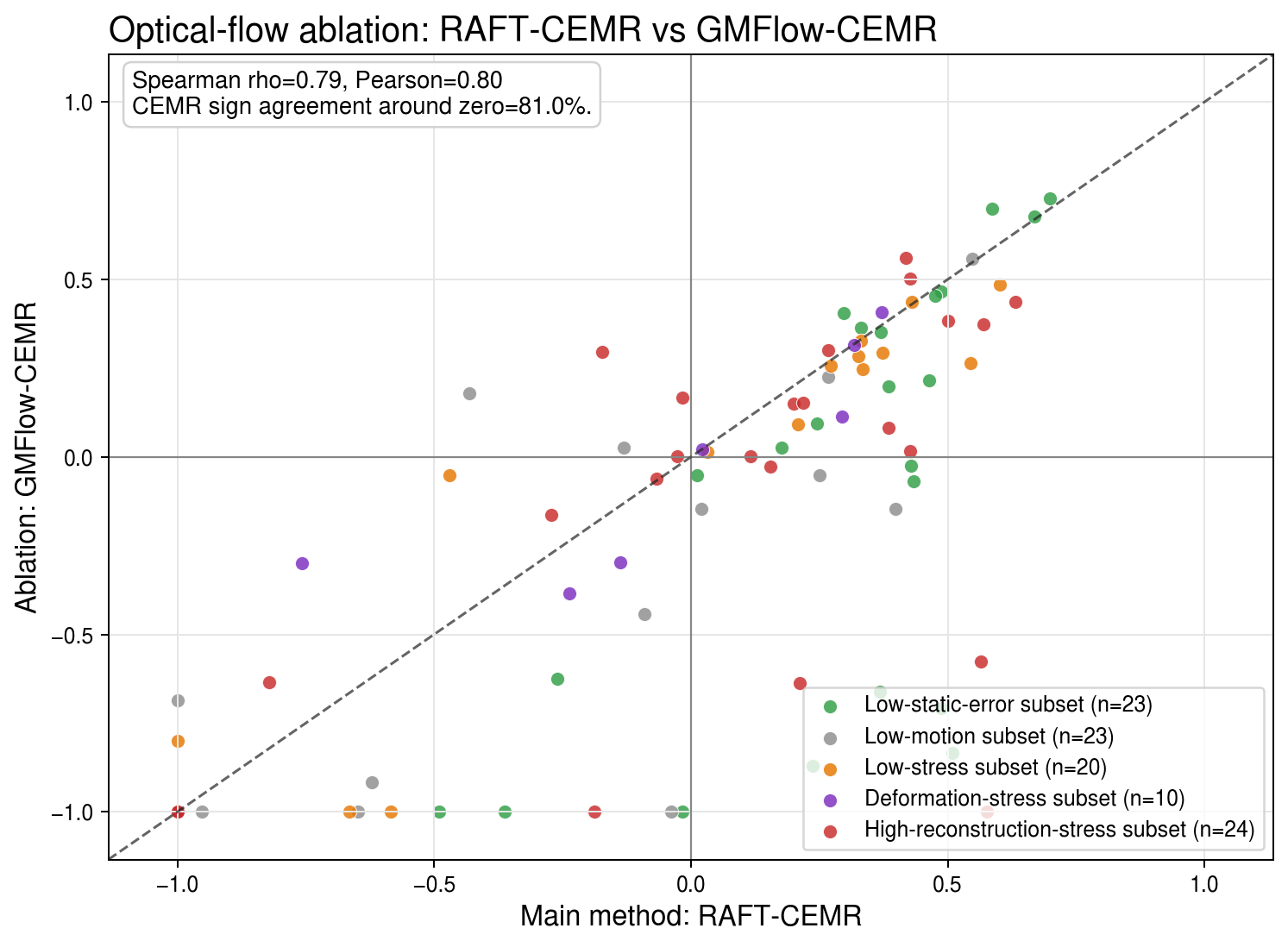}
\caption{Optical-flow ablation visualized for the CEMR diagnostic. Points compare RAFT-CEMR and GMFlow-CEMR on the 100-clip backend-bias subset, colored by diagnostic sampling stratum. The positive rank trend supports CEMR as more than a RAFT-only artifact, while off-diagonal points show why exact CEMR values remain backend-sensitive.}
\label{fig:gmflow-cemr-ablation}
\end{figure}

No single raw metric summarizes the continuous reconstruction profile on the fair-format four-seed panel.
\geco is useful precisely as a contrastive failure analysis: it sometimes agrees with high appearance or gap stress, but its sample-level correlation with CEMR is near zero and many clips with low \geco fused inconsistency still have poor reconstruction diagnostics.
This supports the same-camera reconstruction audit as a complement to local-geometry scoring.
CEMR adds complementary information: Table~\ref{tab:failure-breakdown} shows that high-stress clips can also raise MedianGS flow, MedianGS appearance residuals, and local-geometry diagnostics, often in combination.

\paragraph{Relaxed deformation-capacity stress test.}
The high-gap boundary diagnostic asks whether extra flexible capacity can lower DeformableGS rendering error while the static reconstruction gap remains large.
To test this, we rerun a random stratified 100-clip high-stress subset drawn from the four-seed fair-format panel: 80 High-reconstruction-stress samples and 20 Low-stress samples.
The relaxed setting reuses the same VGGT-estimated camera scenes but increases DeformableGS capacity and optimization budget to 15k iterations, earlier deformation warm-up, higher geometry and deformation learning rates, and densification through 15k.
This capacity stress test increases flexible reconstruction capacity; the DeformableGS training loss used here has no explicit deformation-energy penalty, and deformation energy remains a post-hoc diagnostic.
Table~\ref{tab:deformation-capacity-stress} shows that 14/100 clips become activated-like.
The effect is concentrated in High-reconstruction-stress samples (13/80) and especially WAN high-gap cases (3/10), while Low-stress samples rarely convert (1/20).
Mean residual drops vary across clips: relaxation exposes a measurable subset where extra flexible capacity lowers DeformableGS residual while a high static reconstruction gap remains.
These results support high-gap flexible-fit behavior as a real boundary mode in the reconstruction profile.

\begin{table}[!htbp]
\centering
\scriptsize
\resizebox{\linewidth}{!}{%
\begin{tabular}{lrrrrrrr}
\toprule
Group & $n$ & DefGS residual drop$\uparrow$ & MedianGS residual drop$\uparrow$ & Relaxed gap$\uparrow$ & Deform energy ratio & Activated clips & Activated rate$\uparrow$ \\
\midrule
Overall active diagnostic subset & 100 & -0.022 & -0.073 & 0.121 & 1.918 & 14 & 14.0 \\
Low-stress subset & 20 & -0.074 & -0.098 & 0.097 & 2.710 & 1 & 5.0 \\
High-reconstruction-stress subset & 80 & -0.009 & -0.066 & 0.126 & 1.785 & 13 & 16.2 \\
Other high-gap stress cases & 90 & -0.033 & -0.080 & 0.114 & 1.965 & 11 & 12.2 \\
WAN high-gap stress cases & 10 & 0.075 & -0.006 & 0.180 & 1.575 & 3 & 30.0 \\
\bottomrule
\end{tabular}}
\caption{Relaxed deformation-capacity stress test on a random stratified 100-clip high-stress subset from the fair-format four-seed panel. The relaxed setting reuses the same VGGT-estimated camera scenes but increases DeformableGS capacity and optimization budget. DefGS and MedianGS residual drops are baseline LPIPS minus relaxed LPIPS, so larger positive values indicate lower residual after relaxation. Relaxed gap is the relaxed MedianGS-minus-DeformableGS LPIPS gap. Activated clips have a DeformableGS LPIPS drop of at least 0.05, relaxed static reconstruction gap at least 0.15, and lower relaxed DeformableGS than MedianGS residual. Negative mean drops show that relaxed capacity does not uniformly improve residuals.}
\label{tab:deformation-capacity-stress}
\end{table}

\begin{figure}[!htbp]
\centering
\includegraphics[width=\linewidth]{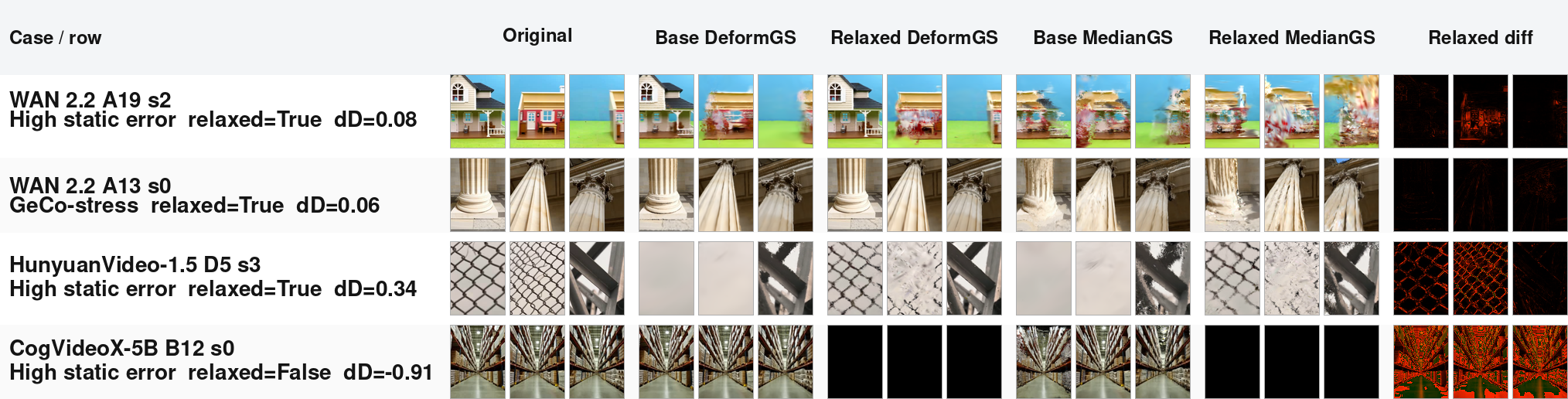}
\caption{Reconstruction renderings for the relaxed deformation-capacity stress test. Rows show selected clips from Table~\ref{tab:deformation-capacity-stress}; columns compare generated observations, the primary 10k DeformableGS fit, the relaxed 15k DeformableGS fit, the primary MedianGS static reconstruction, the relaxed MedianGS static reconstruction, and absolute differences to the relaxed DeformableGS render over three sampled frames. Activated rows illustrate cases where extra flexible capacity lowers DeformableGS residual while a static reconstruction gap remains. The non-activated row shows a high-stress clip that remains difficult under relaxation.}
\label{fig:deformation-capacity-renders}
\end{figure}

\begin{table}[!htbp]
\centering
\scriptsize
\resizebox{\linewidth}{!}{%
\begin{tabular}{lrrrrrrl}
\toprule
Model & Active fail & Med. LPIPS & Med. flow & CEMR & GeCo & Deform. & Dominant reading \\
\midrule
LTX-2.3 HQ & 291 & 262 & 262 & 112 & 61 & 89 & Median LPIPS, Median flow \\
Hunyuan-1.5 & 291 & 260 & 283 & 105 & 49 & 117 & Median flow, Median LPIPS \\
LTX-2.3 dev & 282 & 272 & 244 & 127 & 44 & 98 & Median LPIPS, Median flow \\
WAN 2.2 & 238 & 209 & 215 & 70 & 59 & 103 & Median flow, Median LPIPS \\
CogVideoX-5B & 178 & 134 & 146 & 80 & 36 & 57 & Median flow, Median LPIPS \\
CogVideoX-2B & 137 & 98 & 103 & 66 & 23 & 34 & Median flow, Median LPIPS \\
LTX-Video & 130 & 72 & 90 & 82 & 17 & 21 & Median flow, CEMR \\
Open-Sora & 96 & 74 & 66 & 39 & 23 & 18 & Median LPIPS, Median flow \\
Hunyuan & 56 & 41 & 44 & 33 & 11 & 33 & Median flow, Median LPIPS \\
SkyReels & 52 & 28 & 34 & 33 & 12 & 13 & Median flow, CEMR \\
CogVideoX1.5 & 41 & 29 & 22 & 31 & 2 & 8 & CEMR, Median LPIPS \\
MAGI-1 & 39 & 6 & 10 & 39 & 4 & 1 & CEMR, Median flow \\
\bottomrule
\end{tabular}}
\caption{Non-exclusive diagnostic-layer breakdown for high-stress clips. Counts are clips among generated samples that activate each diagnostic column; columns are not mutually exclusive.}
\label{tab:failure-breakdown}
\end{table}

\begin{table}[!htbp]
\centering
\scriptsize
\resizebox{0.7\linewidth}{!}{%
\begin{tabular}{llr}
\toprule
Check & Metric & Result \\
\midrule
Fair-format four-seed audit & metric completion & $3840/3840$ \\
Fair-format four-seed audit & CEMR completion & $3840/3840$ \\
Seed0 native vs fair-format & MedianGS LPIPS Spearman & $\rho=0.769$ ($n=938$) \\
Seed0 native vs fair-format & VGGT motion-context Spearman & $\rho=0.961$ ($n=960$) \\
Held-out pilot & successful clips & $51/52$ \\
Full-fit vs held-out & MedianGS LPIPS Spearman & $\rho=0.992$ ($n=50$) \\
\bottomrule
\end{tabular}}
\caption{Protocol sanity checks, not benchmark outputs. The held-out values use the actual interleaved held-out pilot metrics, not manifest fields copied from full-fit runs.}
\label{tab:protocol-checks}
\end{table}

\begin{table}[!htbp]
\centering
\scriptsize
\resizebox{\linewidth}{!}{%
\begin{tabular}{lrrrrr}
\toprule
Fair-format diagnostic group & Native low-motion & Native low-static-error & Native low-stress & Native deformation-stress & Native high-stress \\
\midrule
Under-acq. & 212 & 18 & 88 & 2 & 116 \\
Low-static-error & 4 & 1 & 6 & 1 & 33 \\
Low-stress & 0 & 1 & 150 & 0 & 41 \\
Deformation-stress & 0 & 0 & 0 & 0 & 0 \\
High-reconstruction-stress & 4 & 9 & 98 & 3 & 160 \\
\bottomrule
\end{tabular}}
\caption{Seed-0 native-format reversal sanity check on the 12-model overlap with completed static metrics ($947/960$ clips). Rows use fair-format diagnostic strata, and columns use a native-format static-reconstruction proxy without CEMR because native-format CEMR is not recomputed. The table reports format sensitivity for the continuous reconstruction profile.}
\label{tab:native-format-reversal}
\end{table}

\begin{figure}[!htbp]
\centering
\includegraphics[width=\linewidth]{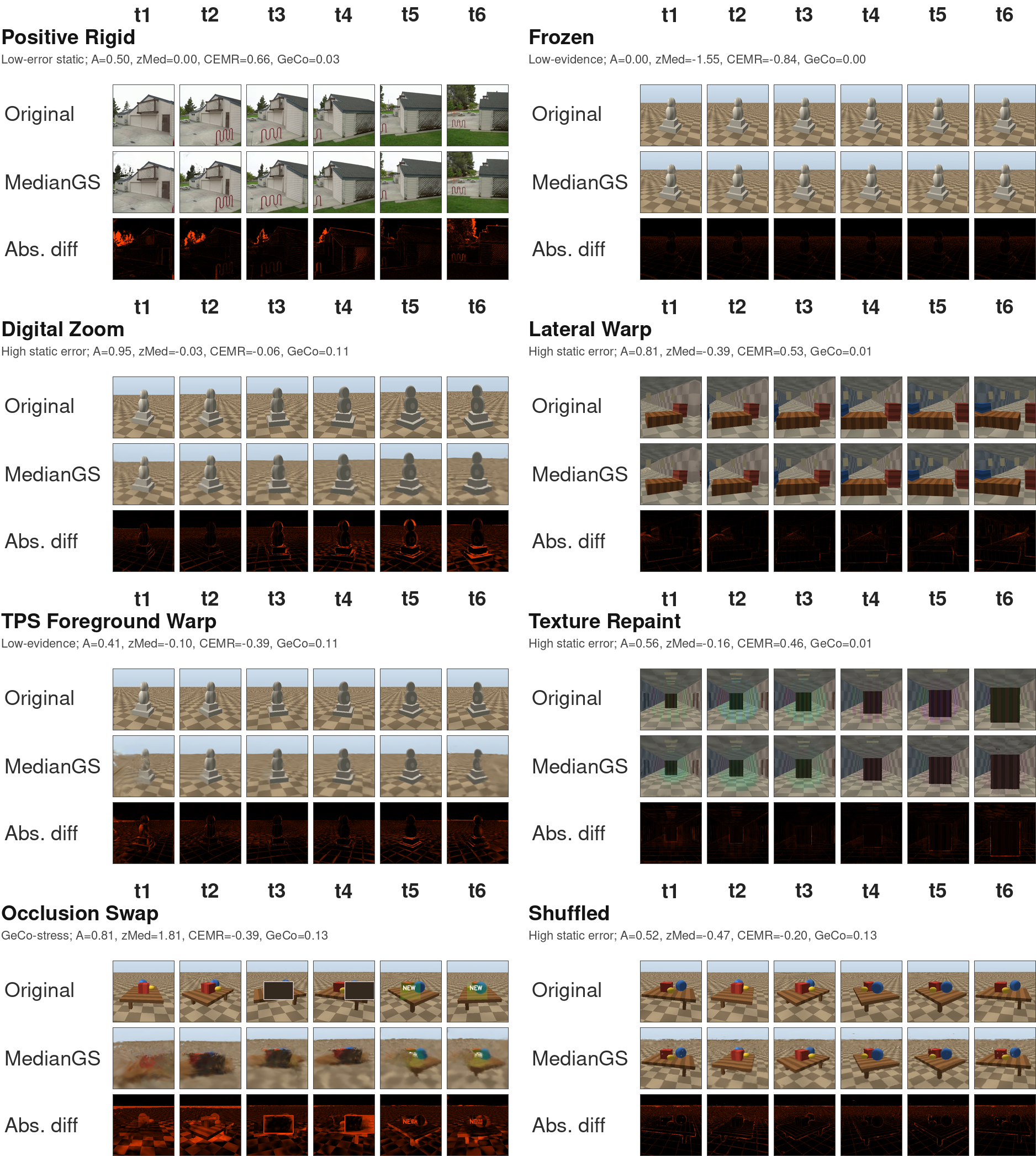}
\caption{ControlBench qualitative examples. Panels show original observations, MedianGS same-camera renders, and absolute differences over six sampled frames for the aggregated ControlBench families, plus two qualitative-only stressors, occlusion swap and shuffled, that are not included in Table~\ref{tab:controlbench-stress-profile}. Panel text gives acquisition and reconstruction diagnostics such as apparent motion, MedianGS rendering error, CEMR, and \geco{} fused inconsistency.}
\label{fig:app-controlbench-qualitative}
\end{figure}

\begin{figure}[!htbp]
\centering
\includegraphics[width=0.60\linewidth]{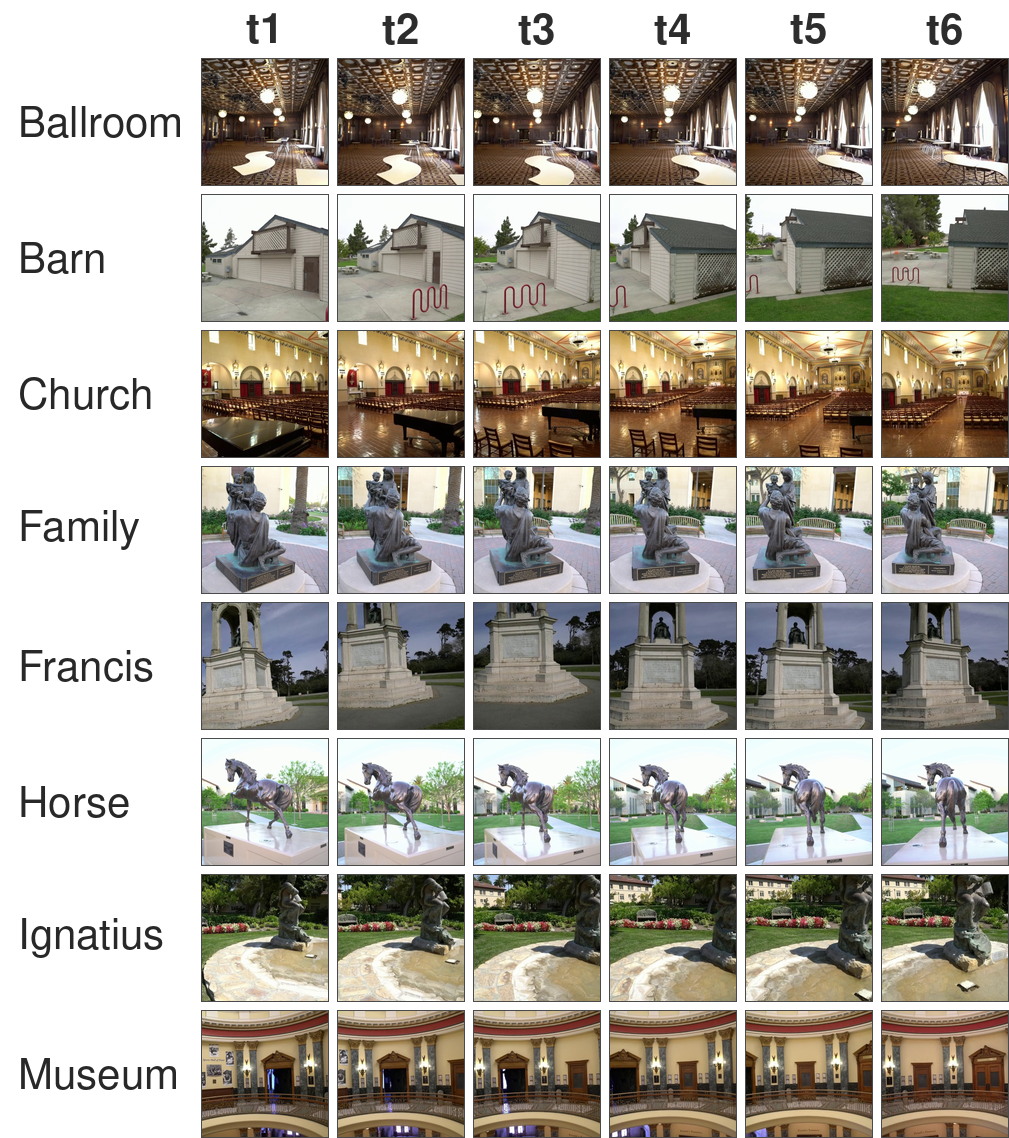}
\caption{Representative public positive controls. Each row is a static scene sampled along its acquisition trajectory. These clips provide real static-world camera evidence for interpreting the reconstruction diagnostics used in ControlBench and the larger stress-test suite.}
\label{fig:app-public-positive-overview}
\end{figure}

\paragraph{ControlBench provenance and qualitative checks.}
ControlBench uses public real static acquisitions from Tanks and Temples scenes~\citep{knapitsch2017tanks} distributed with InstantSplat resources~\citep{fan2024instantsplat}, five deterministic procedural static scenes, and real-derived transformations. All clips are normalized to 49 frames, 8 fps, and $256\times256$ resolution. Figure~\ref{fig:app-controlbench-qualitative} shows representative qualitative examples for the aggregated ControlBench families plus two qualitative-only stressors, occlusion swap and shuffled, which are not included in Table~\ref{tab:controlbench-stress-profile}. Figure~\ref{fig:app-public-positive-overview} shows compact representative public-positive scene samples, while Table~\ref{tab:controlbench-stress-profile} summarizes the larger 1000 public-real / 500 transformed-control stress-test profile.

\paragraph{Prompt categories and descriptor definitions.}
Full prompt text is provided for the retained reconstruction gallery below. The main stress groups are object-centric orbits, indoor traverses with repeated texture and occlusion, outdoor large-scale parallax, and appearance-stress prompts with reflection or repeated high-frequency texture. Video evidence descriptors are context variables: visual motion combines frame differences, optical flow, and dynamic-degree cues; texture/detail combines sharpness, edge density, entropy, and contrast; richness adds color and imaging cues; stability summarizes flicker and temporal smoothness proxies.

\FloatBarrier

\section{Representative Reconstruction Visualizations}
\label{app:recon-visualizations}

\begin{figure}[!htbp]
\centering
\includegraphics[width=0.82\linewidth]{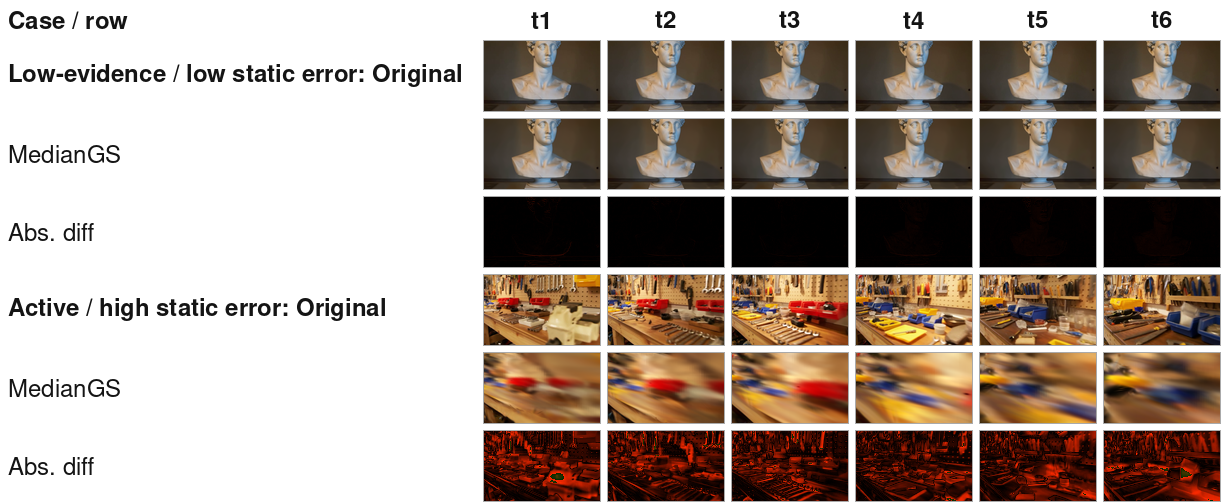}
\caption{Simplified easy-versus-hard acquisition case. Each case shows original frames, MedianGS same-camera renders, and absolute differences. Low-motion clips can be easy to reconstruct, while high-motion/high-richness clips can be visually strong but difficult acquisitions for the current backend. The reconstruction rows are blurry because the backend fails to explain the observations, not because the generated video itself is blurry.}
\label{fig:easy-hard}
\end{figure}

Figure~\ref{fig:easy-hard} provides the compact easy-versus-hard case, and Figure~\ref{fig:app-diagnostics} adds reconstruction diagnostics. The visuals show why acquisition profiles are needed: low-motion clips can be easy to reconstruct without providing much 3D evidence, while active and visually rich clips can be hard acquisitions for the current backend.

\begin{figure}[!htbp]
\centering
\includegraphics[width=\linewidth,height=0.78\textheight,keepaspectratio]{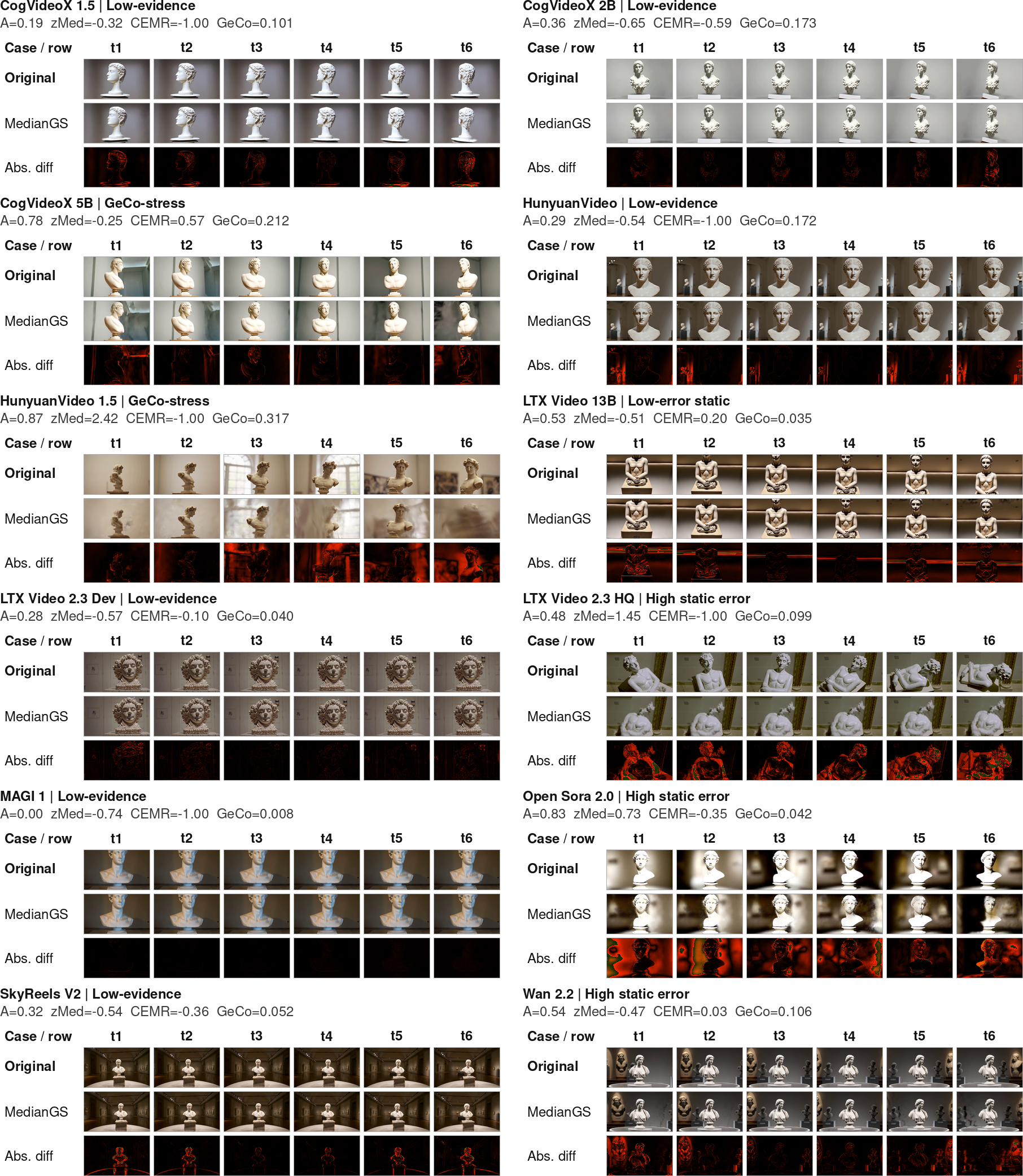}
\caption{Prompt A1: Marble bust orbit, seed 0. Each model panel shows generated observations, MedianGS same-camera renders, and absolute differences over six sampled frames. This object-centric orbit illustrates why low reconstruction residual alone is ambiguous: several stable-looking clips are low-evidence, while more active clips expose static-reconstruction or motion-explanation failures. This appendix retains compact panels; full-resolution artifacts will be released with the benchmark package when licensing permits.}
\label{fig:app-a1-gallery}
\end{figure}

\begin{figure}[!htbp]
\centering
\includegraphics[width=\linewidth,height=0.78\textheight,keepaspectratio]{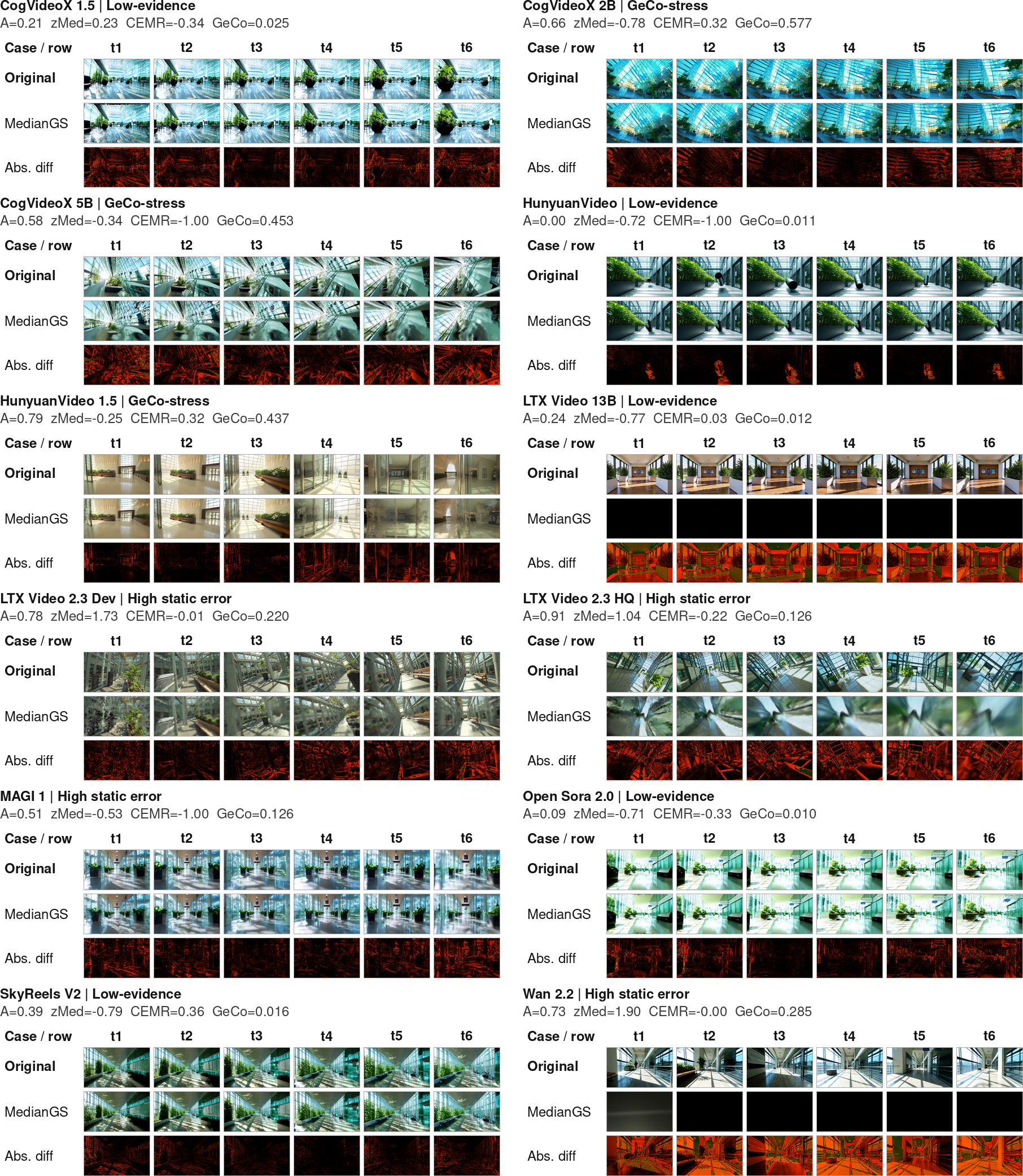}
\caption{Prompt B16: Glass atrium barrel roll, seed 0. Each model panel shows generated observations, MedianGS same-camera renders, and absolute differences over six sampled frames. This indoor roll prompt stresses camera estimation, reflections, and repeated glass structure; the gallery shows that many active-looking clips still produce large static-rendering differences. This appendix retains compact panels; full-resolution artifacts will be released with the benchmark package when licensing permits.}
\label{fig:app-b16-gallery}
\end{figure}

\begin{figure}[!htbp]
\centering
\includegraphics[width=0.95\linewidth]{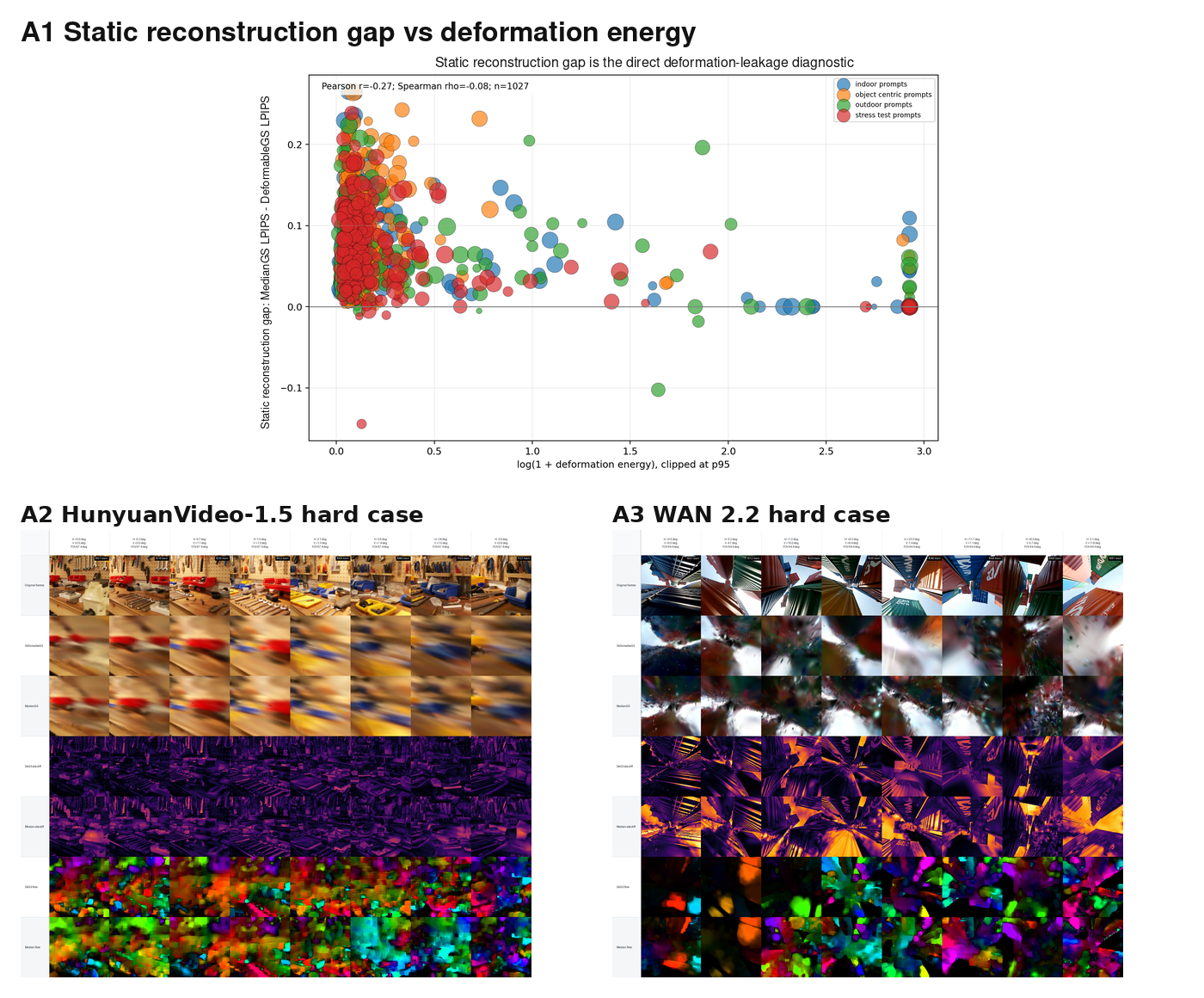}
\caption{Compact appendix diagnostics. Top: static reconstruction gap versus deformation energy, showing that aggregate gaps are often small even when both MedianGS and DeformableGS residuals are high. Bottom: representative hard cases for HunyuanVideo-1.5 and WAN 2.2, where rich active observations are difficult for the current backend to explain as stable static scenes.}
\label{fig:app-diagnostics}
\end{figure}

\paragraph{Prompt-level galleries.}
Figures~\ref{fig:app-a1-gallery} and~\ref{fig:app-b16-gallery} show the two prompt-level galleries retained in this supplementary appendix.
Each panel shows six sampled generated observations, the corresponding MedianGS same-camera renders, and absolute differences for one model configuration.
These examples are not additional rankings; they provide visual evidence for the acquisition profiles reported in the main tables.

\paragraph{Prompt A1: Marble bust orbit.}

\noindent Full prompt: \emph{In a serene gallery, a slow, deliberate 360-degree orbit captures a masterfully carved marble bust, poised elegantly on a polished pedestal. The bust, depicting a serene figure with intricate details, remains perfectly centered as the camera glides smoothly around it. The lighting casts soft shadows, accentuating the delicate features and the smooth texture of the marble. The exposure and white balance are meticulously set, ensuring the bust's timeless beauty is captured in pristine clarity. The focus remains sharp, highlighting every curve and chisel mark, while the background fades into a gentle blur, emphasizing the sculpture's artistry. The only motion is the camera's graceful orbit, creating a mesmerizing, uninterrupted view of this exquisite masterpiece.}

\paragraph{Prompt B16: Glass atrium barrel roll.}

\noindent Full prompt: \emph{The camera embarks on a dynamic barrel-roll journey through a luminous glass atrium corridor, where sunlight streams through expansive windows, casting intricate patterns on the polished floor. As the camera gracefully rotates 180 degrees, the corridor's features remain steadfast: sleek planters brimming with verdant foliage, modern benches inviting rest, and sleek directory boards offering guidance. The reflections on the glass walls and floor remain undisturbed, creating a mesmerizing kaleidoscope of light and shadow. The camera's fluid motion captures the essence of the space, transforming the atrium into a captivating dance of architecture and light.}

\FloatBarrier

\bibliography{refs}

@inproceedings{huang2024vbench,
  title={Vbench: Comprehensive benchmark suite for video generative models},
  author={Huang, Ziqi and He, Yinan and Yu, Jiashuo and Zhang, Fan and Si, Chenyang and Jiang, Yuming and Zhang, Yuanhan and Wu, Tianxing and Jin, Qingyang and Chanpaisit, Nattapol and others},
  booktitle={Proceedings of the IEEE/CVF Conference on Computer Vision and Pattern Recognition},
  pages={21807--21818},
  year={2024}
}

@article{huang2025vbench++,
  title={Vbench++: Comprehensive and versatile benchmark suite for video generative models},
  author={Huang, Ziqi and Zhang, Fan and Xu, Xiaojie and He, Yinan and Yu, Jiashuo and Dong, Ziyue and Ma, Qianli and Chanpaisit, Nattapol and Si, Chenyang and Jiang, Yuming and others},
  journal={IEEE Transactions on Pattern Analysis and Machine Intelligence},
  year={2025},
  publisher={IEEE}
}

@article{zheng2025vbench,
  title={Vbench-2.0: Advancing video generation benchmark suite for intrinsic faithfulness},
  author={Zheng, Dian and Huang, Ziqi and Liu, Hongbo and Zou, Kai and He, Yinan and Zhang, Fan and Gu, Lulu and Zhang, Yuanhan and He, Jingwen and Zheng, Wei-Shi and others},
  journal={arXiv preprint arXiv:2503.21755},
  year={2025}
}

@inproceedings{sun2025t2v,
  title={T2v-compbench: A comprehensive benchmark for compositional text-to-video generation},
  author={Sun, Kaiyue and Huang, Kaiyi and Liu, Xian and Wu, Yue and Xu, Zihan and Li, Zhenguo and Liu, Xihui},
  booktitle={Proceedings of the Computer Vision and Pattern Recognition Conference},
  pages={8406--8416},
  year={2025}
}

@inproceedings{liu2024evalcrafter,
  title={Evalcrafter: Benchmarking and evaluating large video generation models},
  author={Liu, Yaofang and Cun, Xiaodong and Liu, Xuebo and Wang, Xintao and Zhang, Yong and Chen, Haoxin and Liu, Yang and Zeng, Tieyong and Chan, Raymond and Shan, Ying},
  booktitle={Proceedings of the IEEE/CVF conference on computer vision and pattern recognition},
  pages={22139--22149},
  year={2024}
}

@inproceedings{asim2025met3r,
  title={Met3r: Measuring multi-view consistency in generated images},
  author={Asim, Mohammad and Wewer, Christopher and Wimmer, Thomas and Schiele, Bernt and Lenssen, Jan Eric},
  booktitle={Proceedings of the IEEE/CVF Conference on Computer Vision and Pattern Recognition},
  pages={6034--6044},
  year={2025}
}

@inproceedings{duan2025worldscore,
  title={Worldscore: A unified evaluation benchmark for world generation},
  author={Duan, Haoyi and Yu, Hong-Xing and Chen, Sirui and Fei-Fei, Li and Wu, Jiajun},
  booktitle={Proceedings of the IEEE/CVF International Conference on Computer Vision},
  pages={27713--27724},
  year={2025}
}

@article{gu2025geco,
  title={GeCo: A Differentiable Geometric Consistency Metric for Video Generation}, 
  author={Gu, Leslie and Hur, Junhwa and Herrmann, Charles and Zhan, Fangneng and Zickler, Todd and Sun, Deqing and Pfister, Hanspeter},
  journal={arXiv preprint arXiv:2512.22274},
  year={2025}
}

@article{babu2025dynamiceval,
  title={DynamicEval: Rethinking Evaluation for Dynamic Text-to-Video Synthesis},
  author={Babu, Nithin C and Mahapatra, Aniruddha and Rangwani, Harsh and Soundararajan, Rajiv and Kulkarni, Kuldeep},
  journal={arXiv preprint arXiv:2510.07441},
  year={2025}
}

@article{li2024sora,
  title={Sora generates videos with stunning geometrical consistency},
  author={Li, Xuanyi and Zhou, Daquan and Zhang, Chenxu and Wei, Shaodong and Hou, Qibin and Cheng, Ming-Ming},
  journal={arXiv preprint arXiv:2402.17403},
  year={2024}
}

@inproceedings{wang2024motionctrl,
  title={Motionctrl: A unified and flexible motion controller for video generation},
  author={Wang, Zhouxia and Yuan, Ziyang and Wang, Xintao and Li, Yaowei and Chen, Tianshui and Xia, Menghan and Luo, Ping and Shan, Ying},
  booktitle={ACM SIGGRAPH 2024 Conference Papers},
  pages={1--11},
  year={2024}
}

@article{he2024cameractrl,
  title={Cameractrl: Enabling camera control for text-to-video generation},
  author={He, Hao and Xu, Yinghao and Guo, Yuwei and Wetzstein, Gordon and Dai, Bo and Li, Hongsheng and Yang, Ceyuan},
  journal={arXiv preprint arXiv:2404.02101},
  year={2024}
}

@inproceedings{bahmani2025vd3d,
  title={Vd3d: Taming large video diffusion transformers for 3d camera control},
  author={Bahmani, Sherwin and Skorokhodov, Ivan and Siarohin, Aliaksandr and Menapace, Willi and Qian, Guocheng and Vasilkovsky, Michael and Lee, Hsin-Ying and Wang, Chaoyang and Zou, Jiaxu and Tagliasacchi, Andrea and others},
  booktitle={International Conference on Learning Representations},
  volume={2025},
  pages={66712--66737},
  year={2025}
}

@article{hou2024training,
  title={Training-free camera control for video generation},
  author={Hou, Chen and Chen, Zhibo},
  journal={arXiv preprint arXiv:2406.10126},
  year={2024}
}

@article{kerbl20233d,
  title={3d gaussian splatting for real-time radiance field rendering.},
  author={Kerbl, Bernhard and Kopanas, Georgios and Leimk{\"u}hler, Thomas and Drettakis, George and others},
  journal={ACM Trans. Graph.},
  volume={42},
  number={4},
  pages={139--1},
  year={2023}
}

@inproceedings{yang2024deformable,
  title={Deformable 3d gaussians for high-fidelity monocular dynamic scene reconstruction},
  author={Yang, Ziyi and Gao, Xinyu and Zhou, Wen and Jiao, Shaohui and Zhang, Yuqing and Jin, Xiaogang},
  booktitle={Proceedings of the IEEE/CVF conference on computer vision and pattern recognition},
  pages={20331--20341},
  year={2024}
}

@inproceedings{wang2025vggt,
  title={Vggt: Visual geometry grounded transformer},
  author={Wang, Jianyuan and Chen, Minghao and Karaev, Nikita and Vedaldi, Andrea and Rupprecht, Christian and Novotny, David},
  booktitle={Proceedings of the Computer Vision and Pattern Recognition Conference},
  pages={5294--5306},
  year={2025}
}

@inproceedings{wang2024dust3r,
  title={Dust3r: Geometric 3d vision made easy},
  author={Wang, Shuzhe and Leroy, Vincent and Cabon, Yohann and Chidlovskii, Boris and Revaud, Jerome},
  booktitle={Proceedings of the IEEE/CVF conference on computer vision and pattern recognition},
  pages={20697--20709},
  year={2024}
}

@inproceedings{leroy2024grounding,
  title={Grounding image matching in 3d with mast3r},
  author={Leroy, Vincent and Cabon, Yohann and Revaud, J{\'e}r{\^o}me},
  booktitle={European conference on computer vision},
  pages={71--91},
  year={2024},
  organization={Springer}
}

@article{keetha2025mapanything,
  title={Mapanything: Universal feed-forward metric 3d reconstruction},
  author={Keetha, Nikhil and M{\"u}ller, Norman and Sch{\"o}nberger, Johannes and Porzi, Lorenzo and Zhang, Yuchen and Fischer, Tobias and Knapitsch, Arno and Zauss, Duncan and Weber, Ethan and Antunes, Nelson and others},
  journal={arXiv preprint arXiv:2509.13414},
  year={2025}
}

@article{li2026calibanyview,
  title={CalibAnyView: Beyond Single-View Camera Calibration in the Wild},
  author={Li, Boying and Zhang, Cheng and Chen, Weirong and Cremers, Daniel and Reid, Ian and Rezatofighi, Hamid},
  journal={arXiv preprint arXiv:2605.14615},
  year={2026}
}

@article{liu2026reconx,
  title={Reconx: Reconstruct any scene from sparse views with video diffusion model},
  author={Liu, Fangfu and Sun, Wenqiang and Wang, Hanyang and Wang, Yikai and Sun, Haowen and Ye, Junliang and Zhang, Jun and Duan, Yueqi},
  journal={IEEE Transactions on Image Processing},
  year={2026},
  publisher={IEEE}
}

@inproceedings{zhang2018unreasonable,
  title={The unreasonable effectiveness of deep features as a perceptual metric},
  author={Zhang, Richard and Isola, Phillip and Efros, Alexei A and Shechtman, Eli and Wang, Oliver},
  booktitle={Proceedings of the IEEE conference on computer vision and pattern recognition},
  pages={586--595},
  year={2018}
}

@inproceedings{farneback2003two,
  title={Two-frame motion estimation based on polynomial expansion},
  author={Farneb{\"a}ck, Gunnar},
  booktitle={Scandinavian conference on Image analysis},
  pages={363--370},
  year={2003},
  organization={Springer}
}

@inproceedings{teed2020raft,
  title={Raft: Recurrent all-pairs field transforms for optical flow},
  author={Teed, Zachary and Deng, Jia},
  booktitle={European conference on computer vision},
  pages={402--419},
  year={2020},
  organization={Springer}
}

@inproceedings{xu2022gmflow,
  title={Gmflow: Learning optical flow via global matching},
  author={Xu, Haofei and Zhang, Jing and Cai, Jianfei and Rezatofighi, Hamid and Tao, Dacheng},
  booktitle={Proceedings of the IEEE/CVF conference on computer vision and pattern recognition},
  pages={8121--8130},
  year={2022}
}

@inproceedings{schonberger2016structure,
  title={Structure-from-motion revisited},
  author={Schonberger, Johannes L and Frahm, Jan-Michael},
  booktitle={Proceedings of the IEEE conference on computer vision and pattern recognition},
  pages={4104--4113},
  year={2016}
}

@article{savitzky1964smoothing,
  title={Smoothing and differentiation of data by simplified least squares procedures.},
  author={Savitzky, Abraham and Golay, Marcel JE},
  journal={Analytical chemistry},
  volume={36},
  number={8},
  pages={1627--1639},
  year={1964},
  publisher={ACS Publications}
}

@misc{spearman1961proof,
  title={The proof and measurement of association between two things.},
  author={Spearman, Charles},
  year={1961},
  publisher={Appleton-Century-Crofts}
}

@article{knapitsch2017tanks,
  title={Tanks and temples: Benchmarking large-scale scene reconstruction},
  author={Knapitsch, Arno and Park, Jaesik and Zhou, Qian-Yi and Koltun, Vladlen},
  journal={ACM Transactions on Graphics (ToG)},
  volume={36},
  number={4},
  pages={1--13},
  year={2017},
  publisher={ACM New York, NY, USA}
}

@misc{fan2024instantsplat,
  title={InstantSplat: Unbounded Sparse-view Pose-free Gaussian Splatting in 40 Seconds},
  author={Zhiwen Fan and Wenyan Cong and Kairun Wen and Kevin Wang and Jian Zhang and Xinghao Ding and Danfei Xu and Boris Ivanovic and Marco Pavone and Georgios Pavlakos and Zhangyang Wang and Yue Wang},
  year={2024},
  eprint={2403.20309},
  archivePrefix={arXiv},
  primaryClass={cs.CV}
}

@inproceedings{yang2025cogvideox,
  title={Cogvideox: Text-to-video diffusion models with an expert transformer},
  author={Yang, Zhuoyi and Teng, Jiayan and Zheng, Wendi and Ding, Ming and Huang, Shiyu and Xu, Jiazheng and Yang, Yuanming and Hong, Wenyi and Zhang, Xiaohan and Feng, Guanyu and others},
  booktitle={International Conference on Learning Representations},
  volume={2025},
  pages={83048--83077},
  year={2025}
}

@article{kong2024hunyuanvideo,
  title={Hunyuanvideo: A systematic framework for large video generative models},
  author={Kong, Weijie and Tian, Qi and Zhang, Zijian and Min, Rox and Dai, Zuozhuo and Zhou, Jin and Xiong, Jiangfeng and Li, Xin and Wu, Bo and Zhang, Jianwei and others},
  journal={arXiv preprint arXiv:2412.03603},
  year={2024}
}

@article{wu2025hunyuanvideo,
  title={Hunyuanvideo 1.5 technical report},
  author={Wu, Bing and Zou, Chang and Li, Changlin and Huang, Duojun and Yang, Fang and Tan, Hao and Peng, Jack and Wu, Jianbing and Xiong, Jiangfeng and Jiang, Jie and others},
  journal={arXiv preprint arXiv:2511.18870},
  year={2025}
}

@article{hacohen2024ltx,
  title={Ltx-video: Realtime video latent diffusion},
  author={HaCohen, Yoav and Chiprut, Nisan and Brazowski, Benny and Shalem, Daniel and Moshe, Dudu and Richardson, Eitan and Levin, Eran and Shiran, Guy and Zabari, Nir and Gordon, Ori and others},
  journal={arXiv preprint arXiv:2501.00103},
  year={2024}
}

@article{hacohen2026ltx,
  title={LTX-2: Efficient Joint Audio-Visual Foundation Model},
  author={HaCohen, Yoav and Brazowski, Benny and Chiprut, Nisan and Bitterman, Yaki and Kvochko, Andrew and Berkowitz, Avishai and Shalem, Daniel and Lifschitz, Daphna and Moshe, Dudu and Porat, Eitan and others},
  journal={arXiv preprint arXiv:2601.03233},
  year={2026}
}

@article{teng2025magi,
  title={Magi-1: Autoregressive video generation at scale},
  author={Teng, Hansi and Jia, Hongyu and Sun, Lei and Li, Lingzhi and Li, Maolin and Tang, Mingqiu and Han, Shuai and Zhang, Tianning and Zhang, WQ and Luo, Weifeng and others},
  journal={arXiv preprint arXiv:2505.13211},
  year={2025}
}

@article{zheng2025open,
  title={Open-sora 2.0: Training a commercial-level video generation model in {\$}200k},
  author={Zheng, Zangwei and Peng, Xiangyu and Lou, Yuxuan and Shen, Chenhui and Young, Tom and Guo, Xinying and Wang, Binluo and Xu, Hang and Liu, Hongxin and Jiang, Mingyan and others},
  journal={arXiv preprint arXiv:2503.09642},
  year={2025}
}

@article{chen2025skyreels,
  title={Skyreels-v2: Infinite-length film generative model},
  author={Chen, Guibin and Lin, Dixuan and Yang, Jiangping and Lin, Chunze and Zhu, Junchen and Fan, Mingyuan and Zhang, Hao and Chen, Sheng and Chen, Zheng and Ma, Chengcheng and others},
  journal={arXiv preprint arXiv:2504.13074},
  year={2025}
}

@article{wan2025wan,
  title={Wan: Open and advanced large-scale video generative models},
  author={Wan, Team and Wang, Ang and Ai, Baole and Wen, Bin and Mao, Chaojie and Xie, Chen-Wei and Chen, Di and Yu, Feiwu and Zhao, Haiming and Yang, Jianxiao and others},
  journal={arXiv preprint arXiv:2503.20314},
  year={2025}
}

@article{kingma2014adam,
  title={Adam: A method for stochastic optimization},
  author={Kingma, Diederik P and Ba, Jimmy},
  journal={arXiv preprint arXiv:1412.6980},
  year={2014}
}

@article{wang2004image,
  title={Image quality assessment: from error visibility to structural similarity},
  author={Wang, Zhou and Bovik, Alan C and Sheikh, Hamid R and Simoncelli, Eero P},
  journal={IEEE transactions on image processing},
  volume={13},
  number={4},
  pages={600--612},
  year={2004},
  publisher={IEEE}
}

@article{bookstein2002principal,
  title={Principal warps: Thin-plate splines and the decomposition of deformations},
  author={Bookstein, Fred L.},
  journal={IEEE Transactions on pattern analysis and machine intelligence},
  volume={11},
  number={6},
  pages={567--585},
  year={2002},
  publisher={IEEE}
}

\end{document}